\documentclass[letterpaper, 10 pt, journal]{ieeeconf}

\ifdefined\pdfsuppressptexinfo
\fi

\IEEEoverridecommandlockouts
\usepackage{amsmath,amssymb,amsfonts}

\usepackage{bbm}
\usepackage{hyperref}
\hypersetup{
    colorlinks=true,
    linkcolor=orange,
    filecolor=orange,      
    urlcolor=orange,
    citecolor=orange,
}
\usepackage[capitalise, nameinlink]{cleveref}

\usepackage{graphicx}
\usepackage[dvipsnames]{xcolor}
\usepackage{colortbl}
\usepackage{multirow}
\usepackage{makecell}
\usepackage{booktabs}
\usepackage[export]{adjustbox}
\usepackage{pifont}
\usepackage[percent]{overpic}
\usepackage{microtype}
\usepackage{cite}

\begin{document}

\definecolor{blue10}{rgb}{0.3, 0.5, 0.8}
\definecolor{myblue}{rgb}{0.41568, 0.69803, 0.83137}
\definecolor{myorange}{rgb}{1.0000, 0.4980, 0.0549}
\definecolor{myyellow}{rgb}{0.95294, 0.78823, 0.03529}
\definecolor{mygreen}{rgb}{0.443137255	0.749019608	0.431372549}
\definecolor{myred}{rgb}{0.8392, 0.1529, 0.1569}
\definecolor{mypurple}{rgb}{0.5804, 0.4039, 0.7412}
\definecolor{mygrey}{rgb}{0.498, 0.498, 0.498}
\definecolor{mycyan}{rgb}{0.082, 0.745, 0.812}
\definecolor{mybrown}{rgb}{0.5490, 0.3373, 0.2941}
\definecolor{mypink}{rgb}{0.8901960784313725, 0.4666666666666667, 0.7607843137254902}

\newcommand{\myline}[1]{\textcolor{#1}{\raisebox{0.5ex}{\rule{0.4cm}{3.5pt}}}}

\newcommand{\xmark}{\ding{55}}%

\title{\LARGE \bf Long-Term Multi-Session 3D Reconstruction\\Under Substantial Appearance Change}

\author{Beverley Gorry \qquad \quad Tobias Fischer \qquad \quad Michael Milford \qquad \quad Alejandro Fontan %
\thanks{All authors are with the QUT Centre for Robotics, School of Electrical Engineering and Robotics, Queensland University of Technology, Brisbane, QLD 4000, Australia. This research was partially supported by funding from ARC Laureate Fellowship FL210100156 to MM and ARC DECRA Fellowship DE240100149 to TF. The authors acknowledge continued support from the Queensland University of Technology (QUT) through the Centre for Robotics. Corresponding author email: {\tt\small beverley.gorry@hdr.qut.edu.au}}%
}

\bstctlcite{IEEEexample:BSTcontrol}

\maketitle
\thispagestyle{empty}
\pagestyle{empty}

\begin{abstract}
Long-term environmental monitoring requires the ability to reconstruct and align 3D models across repeated site visits separated by months or years to enable the detection of structural change and the assessment of ecosystem health. However, existing Structure-from-Motion (SfM) pipelines are generally designed for imagery captured within a limited time window and consequently fail in long-term monitoring scenarios such as repeated coral reef surveys with substantial visual and structural change. In this paper, we show that the primary limitation of current approaches lies in their inability to maintain visual correspondences under substantial temporal appearance change. We address this limitation by enforcing cross-session correspondences within a joint SfM reconstruction. Our approach combines handcrafted and learned visual features to establish correspondences across large temporal gaps, enabling the reconstruction of a single coherent 3D model from imagery captured years apart, where standard SfM fails to connect sessions. To ensure scalability and further improve alignment robustness on large datasets, we restrict expensive learned feature matching to a small set of likely cross-session image pairs identified via visual place recognition, which reduces computational cost while improving the reliability of cross-session correspondence establishment. We evaluate our method on long-term underwater datasets exhibiting significant real-world change, and demonstrate consistent joint reconstruction across sessions in cases where existing methods fail to connect the sessions or align them only coarsely.
\end{abstract}

\section{Introduction}
\label{sec:introduction}

Long-term 3D mapping across repeated site visits enables applications such as structural change detection, ecosystem health assessment, and targeted intervention for autonomous environmental monitoring~\cite{joshi2022highdefunderwatermapping,sauder2024scalable}. Coral reefs are a particularly compelling and challenging domain as they are among the most biodiverse and valuable marine ecosystems, yet are increasingly threatened by climate change and other anthropogenic stressors~\cite{hoegh2010impact,llewellyn2015gettingupclose}. Monitoring these environments relies on repeated visual surveys acquired months or years apart to assess changes in reef condition~\cite{dunbabin2019realtimevisioncoral,boittiaux2023eiffeltower}.

A fundamental prerequisite for long-term monitoring is the ability to reconstruct and align 3D models captured at different points in time. This paper addresses the problem of reconstructing a single coherent 3D model from visual surveys acquired years apart under substantial appearance and structural change. Underwater imaging challenges, such as refraction effects, severe perceptual difficulties including large appearance change over time, visual aliasing, and feature sparsity, exacerbate this problem~\cite{sun2018objectrecogunderwatervid,she2023semihierarchical,song2024turtlmap}. 

\begin{figure}[t]
\centering
\begin{minipage}[c]{0.05\linewidth}
  \scriptsize
  \centering
  \rotatebox{90}{2016}\\[1.25cm]
  \rotatebox{90}{2017}\\[1.25cm]
  \rotatebox{90}{2018}
\end{minipage}%
\hfill
\begin{minipage}[c]{0.95\linewidth}
  \includegraphics[width=\linewidth,trim=0cm 0cm 0cm 0cm,clip]{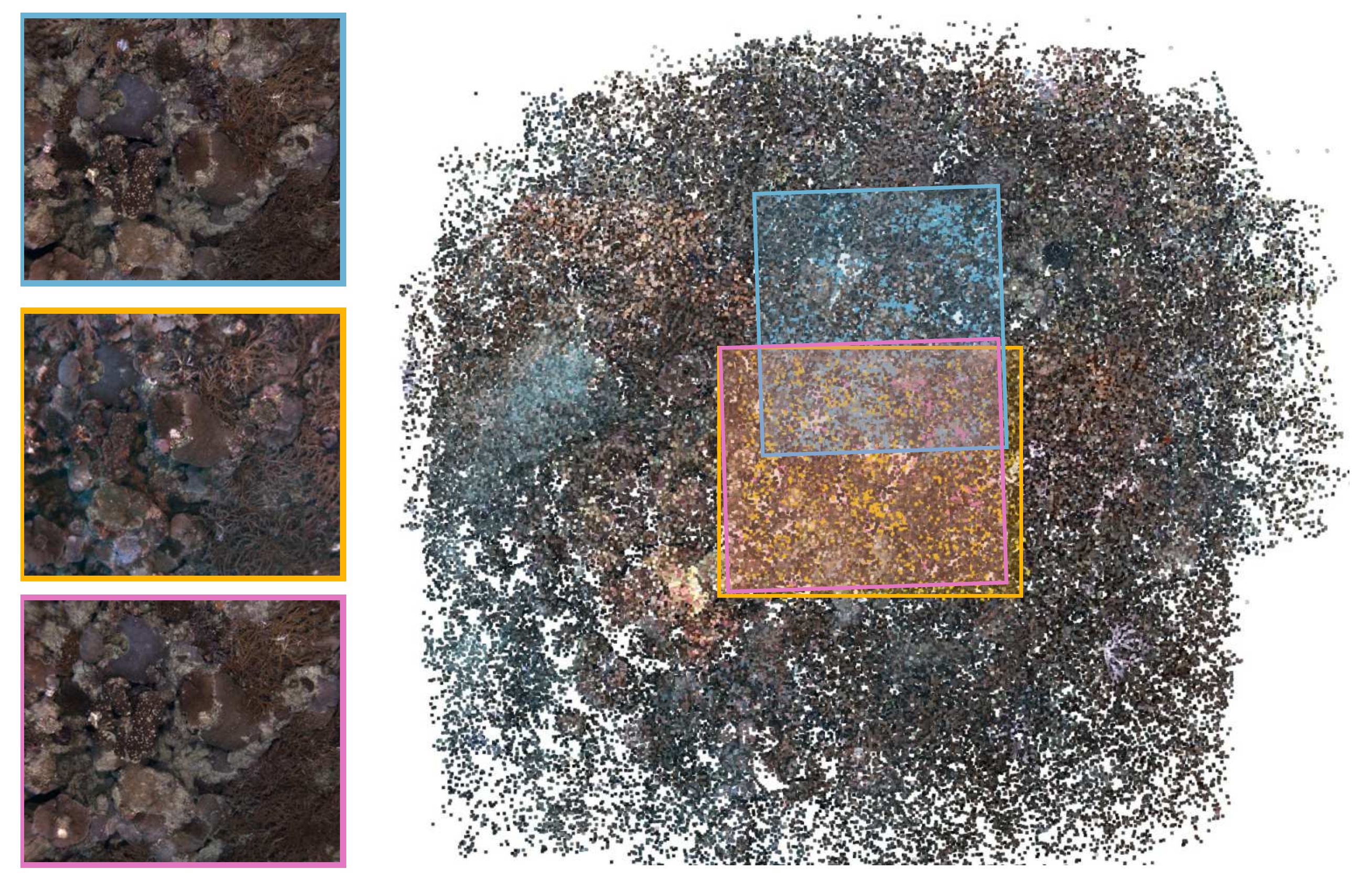}

\end{minipage}
\caption{\textbf{Long-term multi-session 3D reconstruction for coral reef monitoring.} Our method generates a single coherent 3D reconstruction from video-derived images of a coral reef captured over a three-year period (2016--2018). The figure presents the combined 3D point cloud alongside the fields of view of three images acquired from the same location in each year. Despite substantial temporal changes, images of the same area remain aligned with low reprojection error within a shared coordinate frame.}
\vspace*{-0.3cm}
\label{fig:teaser_fig}
\end{figure}
Structure-from-Motion (SfM) and multi-view reconstruction pipelines implicitly assume short capture windows, with limited appearance and structural variation~\cite{boittiaux2023eiffeltower,she2023semihierarchical}. When applied independently to long-term surveys, such pipelines produce reconstructions that lack sufficient cross-session correspondences to support accurate alignment. A common strategy is to reconstruct each session separately and align the resulting point clouds post-hoc~\cite{boittiaux2023eiffeltower,vizzo2023kissicp,pomerleau2015reviewpointcloud}, which assumes each model is already geometrically consistent and shares enough persistent structure across sessions. This strategy is brittle in unstructured natural environments, where the absence of persistent, distinctive geometric landmarks undermines the reliability of post-hoc point cloud alignment and often prevents the formation of a coherent joint reconstruction~\cite{ebadi2023present}.

An alternative to post-hoc alignment is to jointly reconstruct data acquired across multiple time points within a single SfM pipeline. However, existing approaches~\cite{schoenberger2016colmap,keetha2025mapanything,wang2025vggt} also struggle in long-term settings involving large temporal gaps, as they rely on visual feature matching techniques that are insufficiently robust to establish correspondences across substantial appearance change. This typically leads to fragmented reconstructions, the rejection of large portions of the data, and the under-utilization of available imagery.

In this paper, we argue that reliable cross-session correspondence is the key limitation in both independent and joint reconstruction. Post-hoc methods align independently reconstructed sessions without enforcing image-level cross-session constraints during optimization, whereas joint pipelines depend on feature matching that often fails across large temporal changes. We enforce cross-session correspondences directly within the SfM optimization. As illustrated in Figure~\ref{fig:teaser_fig}, this enables video-derived images of the same coral reef area, captured over a three-year period, to be aligned within a shared coordinate frame despite significant temporal changes. 

Specifically, we make the following contributions:
\begin{enumerate}
    \item \textbf{Conceptual contribution: reconstruction under long-term appearance change.} We identify long-term, multi-session 3D reconstruction with substantial appearance and structural change as a practically relevant setting in which SfM pipelines fail. When appearance change is severe and stable geometric landmarks are sparse, aligning independently reconstructed sessions by their geometry alone is insufficient. We show that these sessions can only be aligned accurately when cross-session correspondences are established between them.%

    \item \textbf{Methodological contribution: joint SfM with enforced cross-session correspondences.} We introduce a hybrid feature-based SfM pipeline that enforces cross-session correspondences within a single joint reconstruction by combining handcrafted and learned features. To stay tractable, learned matching is applied only to candidate pairs retrieved by visual place recognition. This yields a unified reconstruction from imagery captured years apart, where handcrafted-only joint SfM fails to connect the sessions.

    \item \textbf{Empirical evaluation on long-term reef datasets.} We evaluate our pipeline on long-term underwater datasets with year-scale appearance change, where it consistently produces coherent joint reconstructions across sessions while existing joint reconstruction and geometry-based alignment methods fail.
\end{enumerate}

\section{Related Work}
\label{sec:related_work}
\textbf{Multi-session and long-term 3D reconstruction.} Long-term and multi-session mapping has been extensively studied in robotics, particularly in lifelong Simultaneous Localization and Mapping (SLAM) and long-term localization. Many approaches explicitly model environment dynamics or persistence over time using occupancy grids or spatio-temporal representations, enabling robots to adapt maps as environments change \cite{tsamis2021towards, krajnik2016persistent}.

A complementary line of research addresses long-term visual localization under appearance change using robust place recognition rather than full metric reconstruction~\cite{gorry2025imagebased}, including recent benchmarks for long-term Visual Place Recognition (VPR) in dynamic benthic environments~\cite{larsen2026sealoc}. Methods based on sequence matching, binary descriptors, or appearance-invariant representations enable reliable relocalization across seasons or illumination conditions, but typically do not produce or maintain consistent multi-session 3D maps~\cite{pomerleau2014long}. More recent systems integrate semantic or object-level representations to improve long-term consistency and scalability, often relying on structured environments, persistent objects, or additional sensing modalities~\cite{adkins2024obvi}.

Despite this substantial body of work, most long-term mapping systems assume that individual sessions can be localized reliably in a shared coordinate frame and that environmental change can be handled through dynamic filtering, semantic abstraction, or map updates. These assumptions become fragile in unstructured natural environments, where appearance and structure may change significantly over time and where independently reconstructed sessions may differ substantially in scale, completeness, and internal consistency. In such settings, post-hoc alignment of independently reconstructed maps can yield superficially overlapping geometry while failing to recover accurate cross-session camera poses.

\textbf{Conventional methods for monitoring corals.} 
Beyond robotics and vision-based mapping, coral reef monitoring has traditionally relied on manual and semi-manual survey protocols. Common approaches include photo quadrats~\cite{obura2019coral,delaval2021status}, in which seafloor images are acquired within a known-area frame along a transect, and video transects~\cite{carleton1995quantitative}, which continuously record the seabed to enable faster data collection. However, both approaches have limitations: photo quadrats are sensitive to quadrat placement, survey protocols, and analyst-dependent interpretation, while video transects lack normalized reference objects and exhibit frame-to-frame overlap and varying viewpoints that complicate benthic cover estimation, increasing annotation effort and uncertainty~\cite{souter2021status}.

\begin{figure*}[t]
\centering
\includegraphics[width=0.9\linewidth]{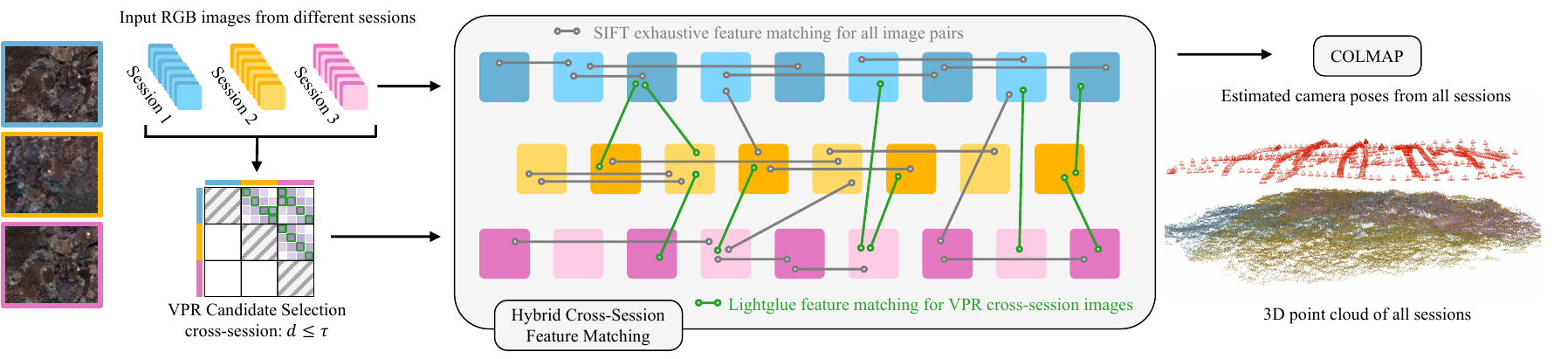}
\vspace*{-0.2cm}
\caption{\textbf{Overview of the proposed reconstruction pipeline.} RGB images acquired from multiple visits to the same area are provided to a feature extraction and matching module, together with a distance matrix generated from Visual Place Recognition. Handcrafted features are matched exhaustively across all image pairs. The rows and columns of the VPR distance matrix list all images session by session. Same-session blocks (hatched) and the lower triangle are excluded, so each pair is used once, and cross-session pairs whose normalized distance is at most $\tau$ (green outlines) are selected as candidates. A learned feature matcher is then applied selectively to these candidates to establish reliable cross-session correspondences, which are enforced directly during joint Structure-from-Motion optimization.}
\vspace*{-0.3cm}
\label{fig:pipeline}
\end{figure*}

\textbf{Computer vision for monitoring corals.} Computer vision has increasingly been applied to coral reef monitoring to automate labor-intensive analysis and enable scalable benthic mapping, including semantic segmentation of live coral from RGB imagery~\cite{sauder2024scalable}. However, underwater imaging poses persistent challenges, such as complex lighting, refraction, caustics, non-linear attenuation, and dynamic scenes, that often confine algorithms to controlled conditions~\cite{sun2018objectrecogunderwatervid,agrafiotis2018underwaterphotogramcaustics}. 

Existing photogrammetry and benthic classification remain sensitive to image quality, requiring curated datasets and substantial computational resources for large-scale reconstruction~\cite{boittiaux2023eiffeltower,bongaerts2021reefscape}. Learning-based benthic classification is also frequently restricted to photo quadrats or orthomosaics and trained with sparse annotations~\cite{chen2021new,yuval2021repeatable}.

Image enhancement is a common preprocessing step for underwater imagery, yet it does not reliably improve feature matching~\cite{summers2025impactunderwaterenhancementfeaturematching}. Learned local features have also been adapted to underwater image degradation, through attenuation-invariant descriptors~\cite{sun2023attenuation}, cross-domain self-supervised training~\cite{cheng2025crossdomain}, or a learned feature extractor within underwater VSLAM~\cite{wang2024ruslam}. These methods target image degradation rather than appearance change between surveys captured years apart.

\textbf{Underwater scene reconstruction.} Many underwater mapping pipelines build on in-air photogrammetry and SfM, most notably COLMAP~\cite{schoenberger2016mvs}, but the underlying assumptions break underwater due to refraction at housing interfaces and non-ideal optics. Existing work addresses this through underwater calibration and refractive imaging~\cite{shortis2015calibration,jordt2016refractive,treibitz2011flat, she2024refractive, LimRefrax}. While these methods improve geometric fidelity, they introduce additional housing assumptions and parameters that limit robustness and ease of deployment in field conditions~\cite{wang2023real}.

\textbf{Feed-forward 3D reconstruction.} Recent models such as MapAnything~\cite{keetha2025mapanything} and VGGT~\cite{wang2025vggt} regress metric scene geometry and camera parameters from one to many views in a single forward pass, but rely on assumptions of static scenes, consistent overlap, and reliable correspondences that do not hold in dynamic underwater environments.

Long-term, multi-session reconstruction remains challenging in terrestrial environments, and it is even more difficult and under-explored in underwater settings, where appearance change, structural disruption, and sensing artifacts make establishing cross-session correspondences especially challenging. %

\section{Methodology}
\label{sec:method}
This section describes the design principles and overall structure of our method for long-term, multi-session 3D reconstruction from repeated visual surveys of natural environments. %

\textbf{Method overview.} The central design principle of our method is that cross-session correspondences must be established and enforced directly within the Structure-from-Motion (SfM) optimization to enable reliable reconstruction under substantial long-term appearance change. Rather than reconstructing each survey visit independently and attempting to align the resulting models post-hoc, we perform a single joint reconstruction that integrates images from all visits into a shared coordinate frame (see Figure~\ref{fig:pipeline}).

This design explicitly targets the primary failure mode of existing pipelines in long-term natural environments: the inability to robustly establish cross-session correspondences when appearance and structure change significantly. By enforcing cross-session constraints during reconstruction, camera poses and 3D structure are jointly optimized across all visits, avoiding the brittleness of post-hoc alignment.

\textbf{Hybrid correspondence strategy.} To support this joint reconstruction, we employ a hybrid correspondence strategy that balances robustness and computational efficiency. Within individual survey visits, where appearance variation is limited, we use fast exhaustive handcrafted feature matching to establish dense and reliable intra-session correspondences. Across different visits, where appearance change is substantial, we selectively apply more robust learned feature matching to establish cross-session correspondences.

Crucially, learned feature matching is not applied exhaustively across all image pairs. Instead, we use Visual Place Recognition (VPR)~\cite{berton2025megalocretrievalplace,larsen2026sealoc} to identify candidate cross-session image pairs that are likely to observe the same physical location. We leverage global descriptors for robust image retrieval under appearance changes and learned local features for precise correspondence estimation~\cite{sattler2018benchmarking6dof,detone2018superpoint}. Learned matching~\cite{lindenberger2023lightglue} is then restricted to these candidates, substantially reducing computational cost while focusing on robustness where it is most needed.

\textbf{Joint reconstruction.} All correspondences, both within and across visits, are provided simultaneously to a single SfM reconstruction pipeline. Camera poses and 3D structure are jointly estimated from this combined correspondence set, yielding a unified reconstruction incorporating imagery from all survey visits in one coordinate frame. We keep the reconstruction backend unchanged, as the limitation we address lies in which correspondences reach the optimizer rather than in the optimization itself. This also enables the approach to be used with existing SfM pipelines.

We emphasize that our approach does not perform pairwise point cloud registration or alignment of independently reconstructed models. Such methods typically assume reliable ground truth, stable geometric landmarks, or independently accurate reconstructions; assumptions that do not hold in unstructured natural environments such as coral reefs.%

\begin{figure}[t]
\centering
\footnotesize

\begin{minipage}[t]{0.48\linewidth}
\centering
\begin{tabular}{ccc}
\multicolumn{3}{c}{{\myline{myblue}} SSK16 {\myline{myyellow}} SSK17 {\myline{mypink}} SSK18}
\end{tabular}

\vspace{0.05cm}
\includegraphics[width=0.95\linewidth]{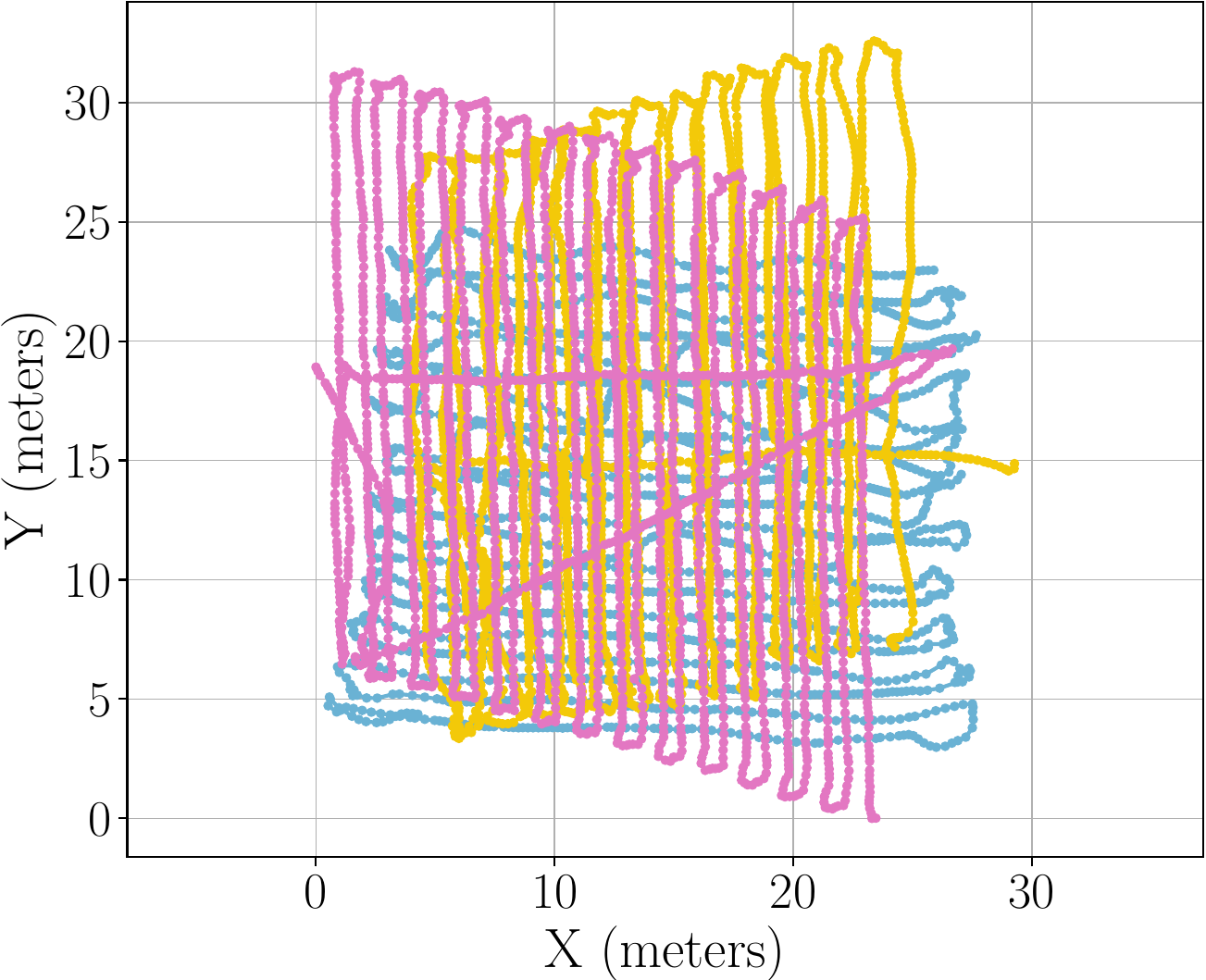}

\vspace{0.1cm}
\includegraphics[width=\linewidth]{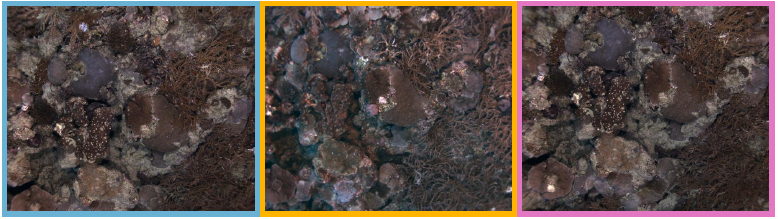}

\vspace{0.05cm}
\end{minipage}
\hfill
\begin{minipage}[t]{0.48\linewidth}
\centering
\begin{tabular}{ccc}
\multicolumn{3}{c}{{\myline{myblue}} EFF16 {\myline{myyellow}} EFF18 {\myline{mypink}} EFF20}
\end{tabular}

\vspace{0.05cm}
\includegraphics[width=0.95\linewidth]{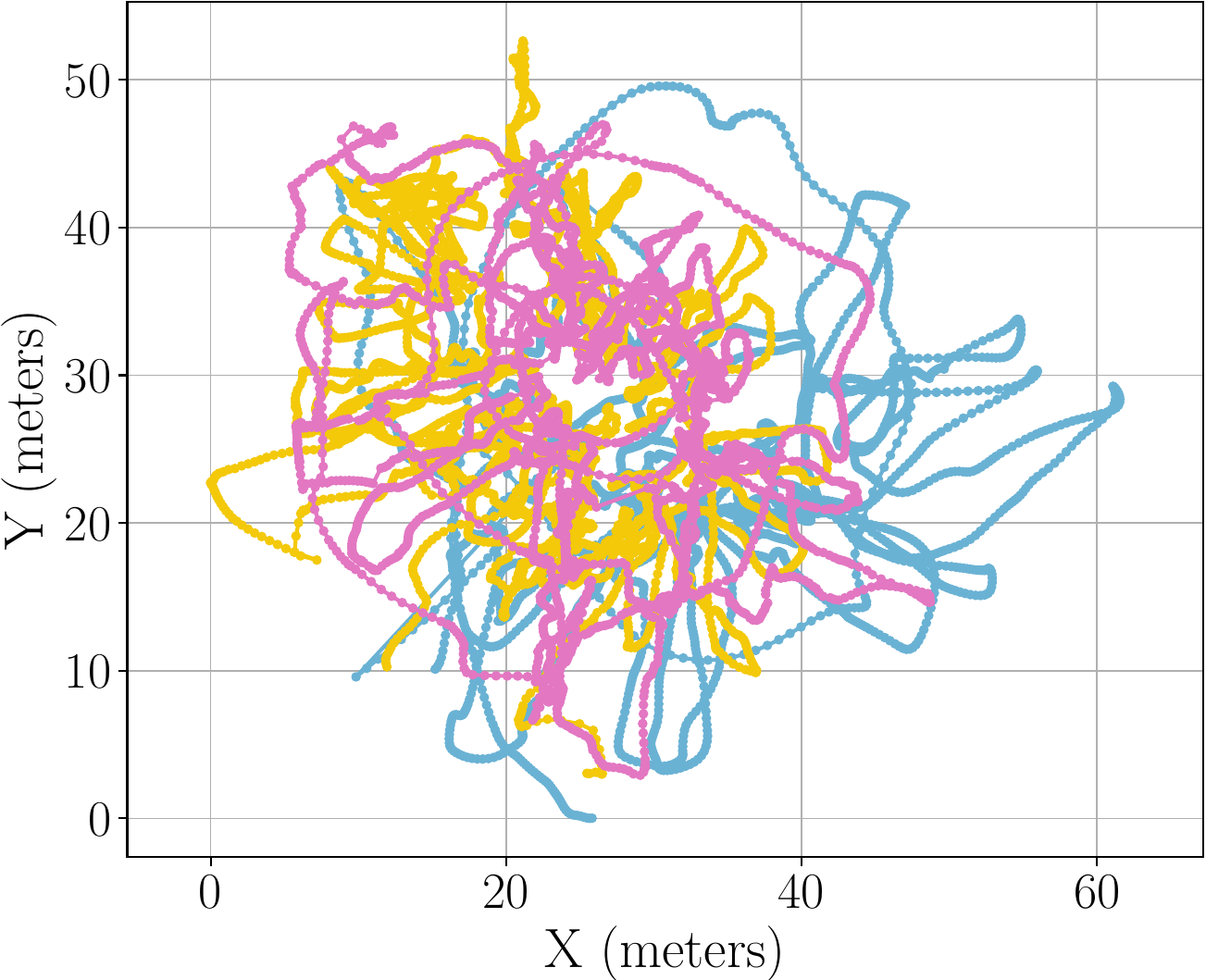}

\vspace{0.25cm}
\includegraphics[width=\linewidth]{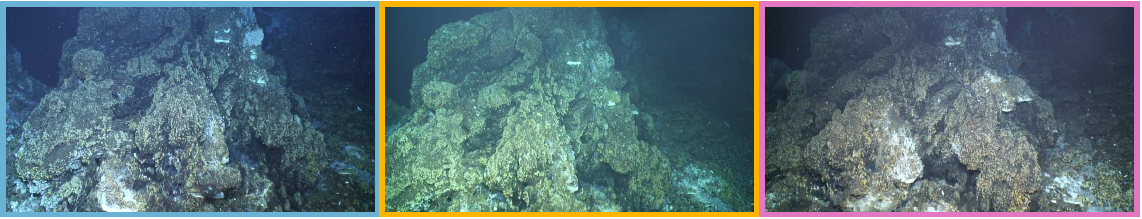}

\vspace{0.05cm}
\end{minipage}

\caption{\textbf{Multi-year underwater vehicle surveys across two long-term underwater datasets.}
\textbf{Left}: Sesoko Island surveys from 2016--2018, showing approximate underwater vehicle navigation trajectories of repeated lawnmower-pattern surveys over the same reef area, with example images captured at the same location from each session. The 2018 session was recorded shortly after Typhoon Trami, which caused substantial appearance variation and structural disruption. \textbf{Right}: Eiffel Tower surveys from 2016, 2018, and 2020, shown in the same layout, with underwater vehicle navigation trajectories on top and images from repeated visits below.}
\label{fig:traj_sesoko_eiffel}
\vspace*{-0.2cm}
\end{figure}

\textbf{Implementation details.} 
We implement the proposed joint reconstruction pipeline on top of the COLMAP SfM backend~\cite{schoenberger2016colmap} without modifying its optimization. We use handcrafted SIFT~\cite{lowe1999sift} features with exhaustive matching across all image pairs to establish dense intra-session correspondences. The VPR method MegaLoc~\cite{berton2025megalocretrievalplace} is used solely to identify candidate cross-session image pairs, selected when their min-max normalized descriptor distance is at most a threshold $\tau$, but does not impose geometric constraints on the reconstruction. We establish cross-session correspondences by selectively applying a learned feature matcher, LightGlue~\cite{lindenberger2023lightglue}, with SuperPoint~\cite{detone2018superpoint} features to these candidate image pairs, using the pretrained weights without tuning for underwater imagery. LightGlue uses its default match confidence threshold with adaptive pruning disabled. A candidate pair contributes learned matches only if it has more than 40 matches, and pairs with fewer than 10 SIFT and learned matches in total are discarded. Learned keypoints and matches are appended to the SIFT keypoints and matches in the COLMAP database.

All intra- and cross-session correspondences are provided simultaneously to the unmodified SfM optimizer, which jointly estimates camera poses and 3D structure. We do not model refraction explicitly. Instead, it is absorbed into the radial distortion of COLMAP's OpenCV camera model, with one set of intrinsics per camera configuration that is shared by all sessions captured with it. Because learned keypoints are matched only across sessions, we initialize COLMAP's incremental mapper from the cross-session image pair with the most verified matches that satisfies COLMAP's initialization criteria. COLMAP's two-view geometric verification retains only matches consistent with the estimated two-view geometry, and the mapper's triangulation and bundle adjustment discard observations with high reprojection error. Unless otherwise stated, we use standard configurations of the underlying SfM and feature matching components, and report all thresholds that materially affect matching behavior.

Due to the downward-facing viewpoint of our imagery, which introduces large and arbitrary rotations about the optical axis, we account for roll in both visual place recognition~\cite{berton2025megalocretrievalplace} and learned feature matching~\cite{lindenberger2023lightglue,detone2018superpoint} by evaluating image rotations of 0$^\circ$, 90$^\circ$, 180$^\circ$, and 270$^\circ$~\cite{fontan2026aerial}. For visual place recognition, the distance between two images is the minimum over the rotations. For learned matching, SuperPoint~\cite{detone2018superpoint} extracts up to 2048 keypoints from each rotation of each image, which are mapped back to the unrotated image. LightGlue~\cite{lindenberger2023lightglue} first matches the first image against each rotation of the second, selects the relative rotation with the most matches, and adds the matches of the three further rotation pairs with the same relative rotation.

Overall, matching remains reliable under substantial appearance change for three reasons: retrieval restricts learned matching to image pairs that likely observe the same location, learned features establish correspondences where handcrafted features do not, and geometric verification and bundle adjustment reject matches that are inconsistent with a single static scene.
\section{Experimental Setup}
\label{sec:experimentalsetup}

\begin{table}[t]
\setlength{\tabcolsep}{3pt}
\newcommand{\rot}[1]{\makebox[1em][l]{\rotatebox{45}{#1}}}
\caption{\textbf{Technical details of the Sesoko and Eiffel Tower datasets, and the subsampled Sesoko evaluation regions.} Images are 608$\times$505 px for Sesoko and 1920$\times$1080 px for Eiffel Tower.}
\label{tab:datasets}
\vspace{-0.25cm}
\centering
\scriptsize
\begin{tabular}{cccccccc}%
\textbf{Dataset} & \rot{\textbf{Year}} & \rot{\textbf{\# Images (k)}} & \rot{\textbf{Dur. (h)}} & \rot{\textbf{Dist. (km)}} & \rot{\textbf{Avg. Depth (m)}} & \rot{\textbf{Seafloor}} & \rot{\textbf{Platform}} \\ \midrule
\multirow{3}{*}{Sesoko} & 2018 & 2.8 & 01:16 & 0.7 & 37 & \multirow{3}{*}{\makecell{Mesophotic \\ coral}} & \multirow{3}{*}{\makecell{SOTON/UTOK \\OPLab AUV}} \\
& 2017 & 3.0 & 02:04 & 0.7 & 36 \\
& 2016 & 2.1 & 00:58 & 0.5 & 38 \\ \midrule
\multirow{3}{*}{\makecell{Eiffel \\Tower}} & 2020 & 3.9 & 03:18 & 0.9 & 1685 & \multirow{3}{*}{\makecell{Hydrothermal \\vent edifice}} & \multirow{3}{*}{\makecell{ EMSO-Azores \\ROV}} \\
& 2018 & 5.4 & 04:34 & 1.0 & 1687 \\
& 2016 & 3.6 & 03:05 & 0.9 & 1686 \\
\bottomrule
\end{tabular}

\vspace{0.4cm}

\begin{tabular}{lccccc}
\textbf{Subset}  & \textbf{R (m)} & \textbf{\# Im. 2016} & \textbf{\# Im. 2017} & \textbf{\# Im. 2018} & \textbf{\# Im. Total} \\
\midrule
ssk-s01 & 2 & 50 & 63 & 52 & 165\\
ssk-s02  & 3 &  83 & 134 & 95 & 312\\
ssk-s03  & 3 &  87 & 129 & 93 & 309\\
ssk-s04 & 3 &  90 & 76 & 110 & 276\\
ssk-s05 & 5 & 230 & 322 & 188 & 740\\
\bottomrule
\end{tabular}
\vspace*{-0.25cm}
\end{table}

\textbf{Datasets.}
We evaluate our approach on two complementary long-term underwater datasets (see Table~\ref{tab:datasets}): the Sesoko coral reef dataset~\cite{nakatani2008auv} and the Eiffel Tower hydrothermal vent dataset~\cite{boittiaux2023eiffeltower}.

We perform experiments on smaller spatial subsets from Sesoko, covering areas approximately 2--5 meters in radius containing 150--750 images. Each subset combines images from all yearly sessions, providing spatial overlap across visits while keeping reconstruction and manual evaluation tractable. We then evaluate scalability and overall reconstruction performance using the complete datasets, retaining every second frame to reduce computational cost. We analyze the sensitivity to the VPR threshold in Section~\ref{sec:ablations} and fix it to values that keep the computational time tractable, $\tau = 0.55$ on the Sesoko subsets, $0.50$ on ssk-full and $0.48$ on eff-full. %

\begin{table*}[t]
\centering
\caption{\textbf{Comparison of alignment accuracy and multi-session connectivity across all methods.} RPE: median reprojection error (pixels); $n$: annotated points evaluated; Sh. Pts.: share of 3D points observed in at least two sessions; Time: mean total runtime (min) over ssk-s01 to ssk-s05. Our method and its two post-hoc variants use the same VPR-selected image pairs and their verified correspondences. \xmark: the method failed to produce a complete point cloud or no annotated point could be evaluated. Best RPE per dataset in \textbf{bold}.}
\resizebox{\textwidth}{!}{%
\setlength{\tabcolsep}{2pt}
\begin{tabular}{lcccccccccccccccccccccc}
& \multicolumn{3}{c}{ssk-s01} & \multicolumn{3}{c}{ssk-s02} & \multicolumn{3}{c}{ssk-s03} & \multicolumn{3}{c}{ssk-s04} & \multicolumn{3}{c}{ssk-s05} & \multicolumn{3}{c}{ssk-full} & \multicolumn{3}{c}{eff-full} & \\
\cmidrule(lr){2-4} \cmidrule(lr){5-7} \cmidrule(lr){8-10} \cmidrule(lr){11-13} \cmidrule(lr){14-16} \cmidrule(lr){17-19} \cmidrule(lr){20-22}
Method & RPE$\downarrow$ & $n\uparrow$ & Sh. Pts.$\uparrow$ & RPE$\downarrow$ & $n\uparrow$ & Sh. Pts.$\uparrow$ & RPE$\downarrow$ & $n\uparrow$ & Sh. Pts.$\uparrow$ & RPE$\downarrow$ & $n\uparrow$ & Sh. Pts.$\uparrow$ & RPE$\downarrow$ & $n\uparrow$ & Sh. Pts.$\uparrow$ & RPE$\downarrow$ & $n\uparrow$ & Sh. Pts.$\uparrow$ & RPE$\downarrow$ & $n\uparrow$ & Sh. Pts.$\uparrow$ & \makecell{Time\\(min)$\downarrow$} \\
\midrule
\multicolumn{23}{l}{\textit{Baselines}} \\
COLMAP & \xmark & 0/59 & 0\% & 2.34 & 37/85 & 0.4\% & 2.36 & 51/103 & 0.3\% & 3.95 & 12/35 & 0.1\% & 5.09 & 128/145 & 0.3\% & 8.10 & 59/98 & 0.1\% & 14.82 & 53/57 & 1\% & 4.3 \\
MapAnything & 334.92 & 56/59 & 0\% & 448.47 & 85/85 & 0\% & 415.15 & 100/103 & 0\% & 76.98 & 35/35 & 0\% & 603.32 & 143/145 & 0\% & \xmark & 0/98 & 0\% & \xmark & 0/57 & 0\% & 1.8 \\
COLMAP + ICP & 371.19 & 56/59 & 0\% & 56.59 & 79/85 & 0\% & 40.15 & 99/103 & 0\% & 404.86 & 34/35 & 0\% & 366.09 & 128/145 & 0\% & 584.83 & 59/98 & 0\% & 108.72 & 52/57 & 0\% & 2.7 \\
COLMAP + BUFFER-X & 54.40 & 56/59 & 0\% & 36.03 & 79/85 & 0\% & 32.79 & 99/103 & 0\% & 68.31 & 34/35 & 0\% & 27.52 & 128/145 & 0\% & 86.75 & 59/98 & 0\% & 354.31 & 52/57 & 0\% & 2.6 \\
\midrule
\multicolumn{23}{l}{\textit{Variants of our method (same VPR pairs and verified correspondences)}} \\
Post-hoc, Sim(3) & 2.40 & 56/59 & 0\% & 3.89 & 79/85 & 0\% & 2.61 & 99/103 & 0\% & 6.60 & 34/35 & 0\% & 4.12 & 128/145 & 0\% & 20.96 & 59/98 & 0\% & 33.69 & 52/57 & 0\% & 10.8 \\
Post-hoc, Sim(3)~+~BA & 1.70 & 59/59 & 23\% & 2.93 & 85/85 & 22\% & 2.31 & 103/103 & 23\% & 5.79 & 35/35 & 12\% & 3.31 & 144/145 & 18\% & 15.00 & 89/98 & 6\% & 28.71 & 55/57 & 7\% & 11.5 \\
Joint (Ours) & \textbf{1.36} & 59/59 & 24\% & \textbf{2.28} & 85/85 & 25\% & \textbf{2.15} & 103/103 & 25\% & \textbf{3.94} & 35/35 & 13\% & \textbf{2.00} & 145/145 & 21\% & \textbf{2.42} & 92/98 & 15\% & \textbf{7.29} & 57/57 & 11\% & 16.7 \\
\end{tabular}}
\label{tab:main_results}
\vspace*{-0.3cm}
\end{table*}

\textbf{Baselines.}
We evaluate our approach against three approaches designed to assess both joint reconstruction robustness under large temporal gaps and post-hoc alignment quality when reconstruction is performed independently per visit.

1) \textbf{Joint 3D Reconstruction}: images from all visits are provided simultaneously to estimate the 3D point cloud and camera poses. We evaluate the SOTA SfM \textbf{COLMAP}~\cite{schoenberger2016colmap} and \textbf{MapAnything}~\cite{keetha2025mapanything}, a recent SOTA end-to-end transformer that regresses metric 3D scene geometry from images. Both are run off-the-shelf, using released weights without fine-tuning and with images as the only input.

2) \textbf{Independent 3D Reconstruction with Post-hoc Alignment.} We estimate the 3D structure of each yearly session separately using COLMAP, then bring each session to metric scale by fitting a similarity transform from its camera centers to the survey's camera positions and applying only its scale. The resulting point clouds are then aligned using \textbf{point-to-point ICP}~\cite{bai2023icp}, and using \textbf{BUFFER-X}~\cite{seo2025buffer}, which combines multi-scale patch-based descriptors with hierarchical inlier search across scales. 

3) \textbf{Post-hoc Alignment with Our Correspondences.} The following two variants of our method use exactly its cross-session evidence, but integrate it \emph{after} reconstruction rather than during it. Beginning with the same independent per-session COLMAP reconstructions as the second group, the matched keypoints are lifted to 3D in their own session's reconstruction. \textbf{Sim(3)} aligns each session to a reference session with a similarity transform estimated from these 3D-3D correspondences (Umeyama's method within RANSAC~\cite{umeyama1991leastsquares,fischler1981ransac}). \textbf{Sim(3) + BA} then triangulates all matches on the aligned sessions and performs bundle adjustment. Comparing these variants with our method separates the contribution of the correspondences from that of their joint integration.

\textbf{Evaluation metrics.}
Because subpixel-accurate ground-truth alignment is unavailable for long-term multi-session reconstructions, we use manual annotations. We select 8--20 spatially distributed image pairs from different sessions. In each pair, an annotator selects distinctive keypoints on structure that is identifiable in both sessions in the query image and marks their corresponding 2D locations in the database image. This yields a fixed set of 35--145 annotated points per dataset that is shared by all methods.

We quantify alignment accuracy by the pixel reprojection error. Additionally, we report the number $n$ of annotated points evaluated by each method and the percentage of 3D points observed in more than one session as measures of cross-session connectivity, as well as the mean total runtime.  All subset experiments were run on an NVIDIA RTX 4090 GPU and Intel Core i7-14700 CPU, while MapAnything and full datasets were run on an NVIDIA H100.

\section{Results}
\label{sec:results}

\textbf{Comparison against baselines.} This experiment evaluates whether existing reconstruction strategies can support long-term, multi-session reconstruction under substantial appearance change. We compare existing joint reconstruction methods, post-hoc alignment of independently reconstructed sessions, and our proposed framework, which enforces cross-session correspondences directly during the SfM optimization.

Alignment accuracy is evaluated using manually annotated cross-session image correspondences (Section~\ref{sec:experimentalsetup}). Figure~\ref{fig:points3d_images} illustrates pixel-level reprojection behavior while Table~\ref{tab:main_results} reports alignment accuracy and cross-session connectivity.
\newcommand{\pairfig}[1]{\includegraphics[width=0.16\textwidth]{figures/pair_01_8312297_to_8315088_#1.png}}
\newcommand{\cloudfig}[1]{\includegraphics[width=0.16\textwidth]{figures/sskall-s01_#1.png}}
\begin{figure*}[t]
\centering
\footnotesize
{\myline{myblue}} SSK16 {\myline{myyellow}} SSK17 {\myline{mypink}} SSK18 {\myline{mygrey}} Multiple sessions\\[0.1cm]
\setlength{\tabcolsep}{1.5pt}
\begin{tabular}{cccccc}
& COLMAP & MapAnything & COLMAP + ICP & COLMAP + BUFFER-X & Ours
\\
& \cloudfig{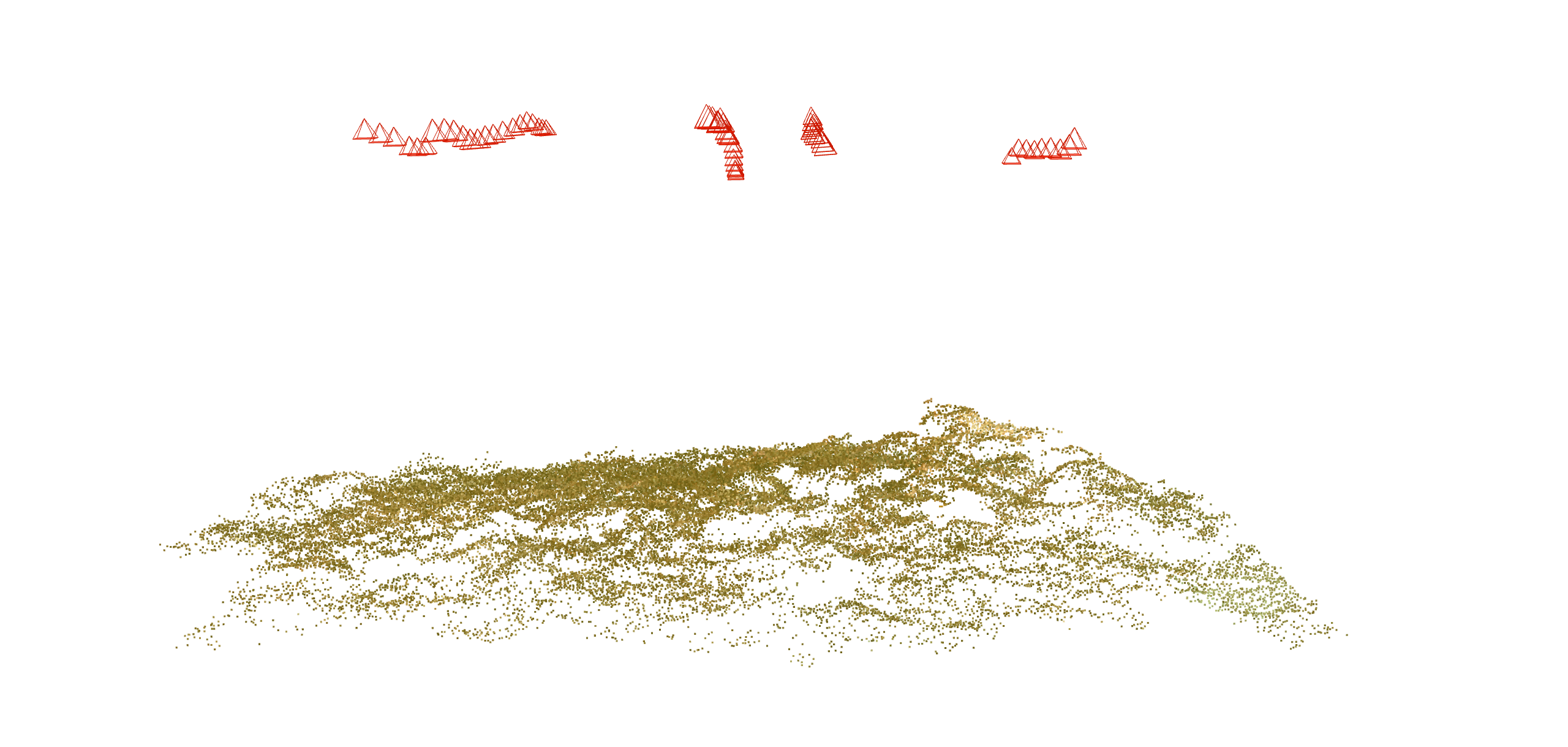} & \cloudfig{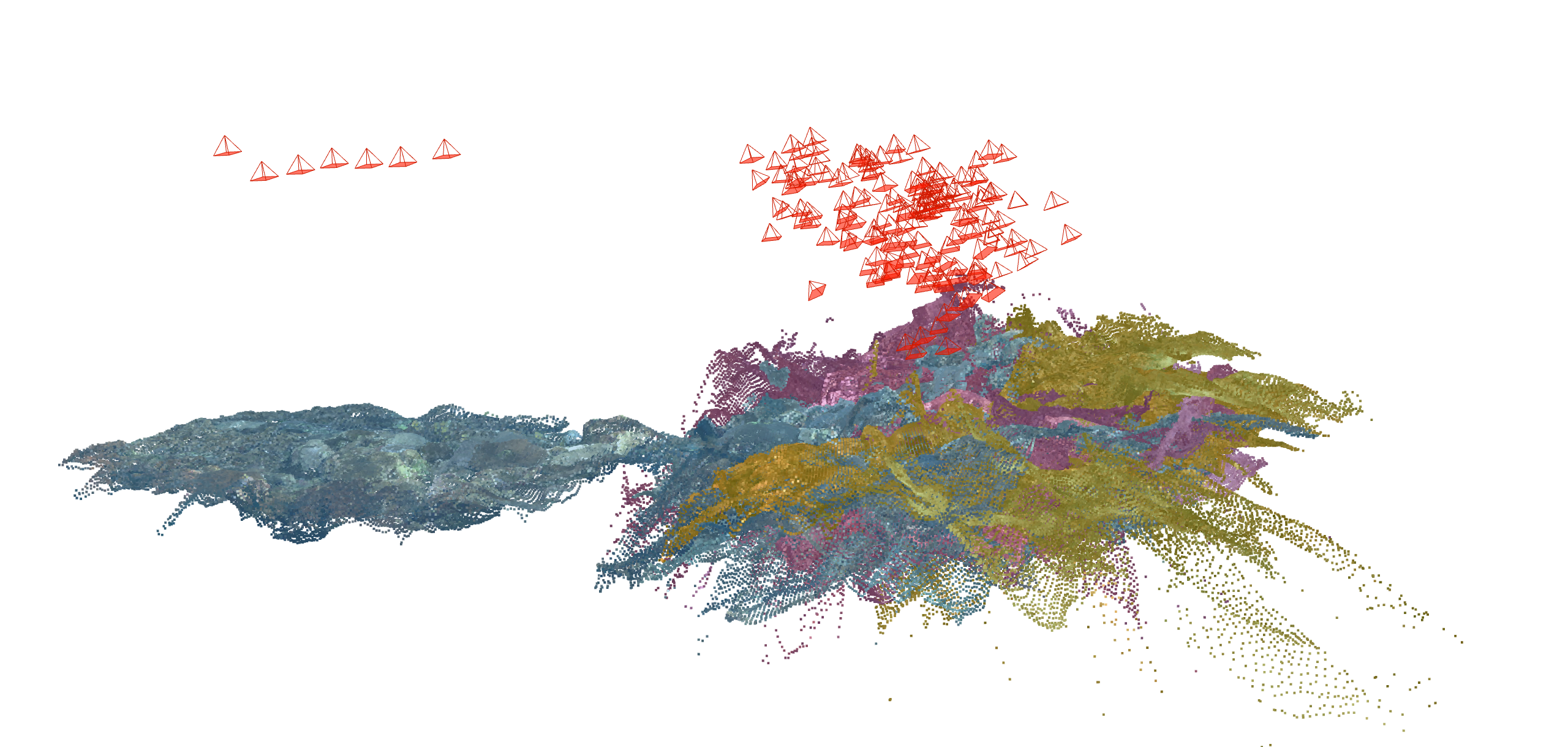} & \cloudfig{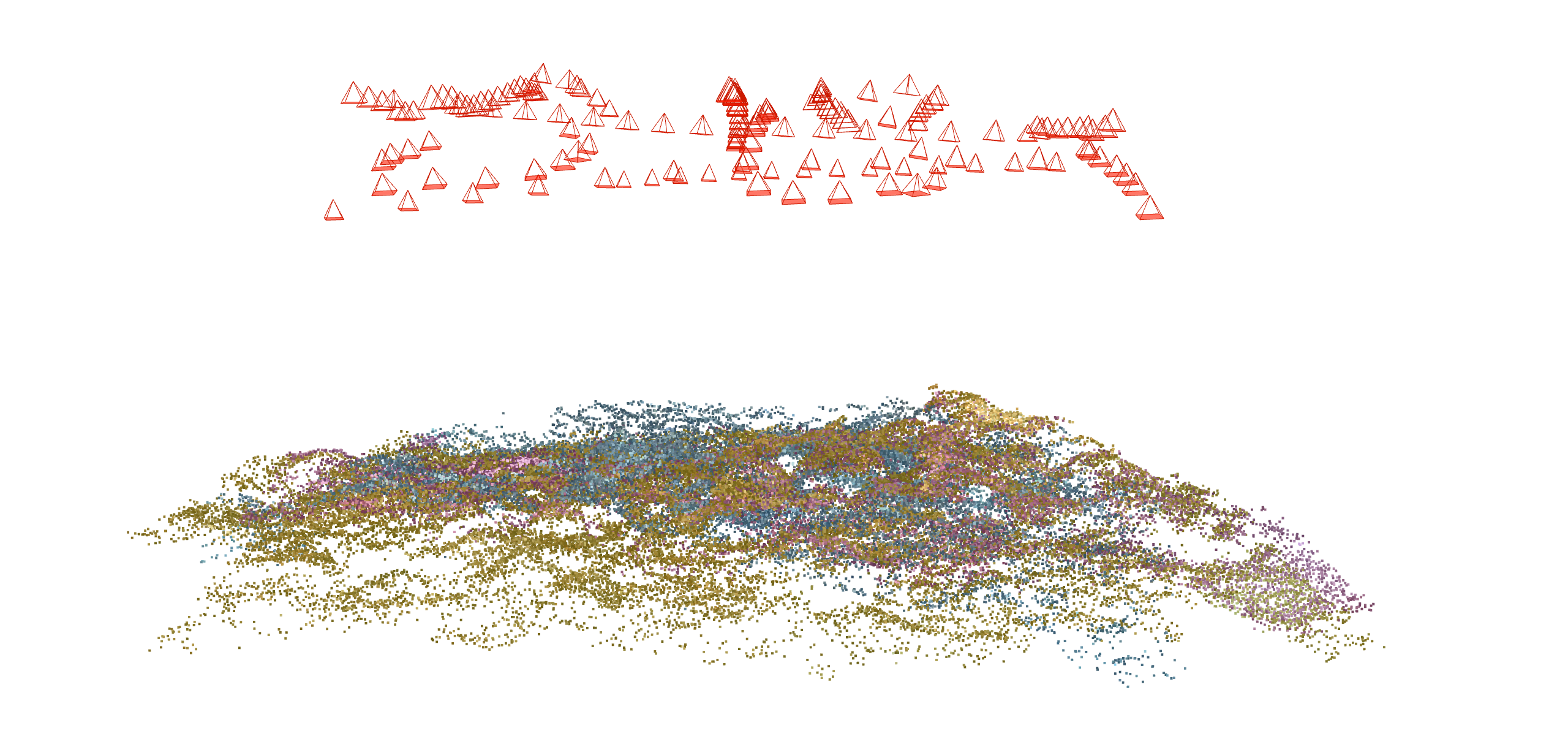} & \cloudfig{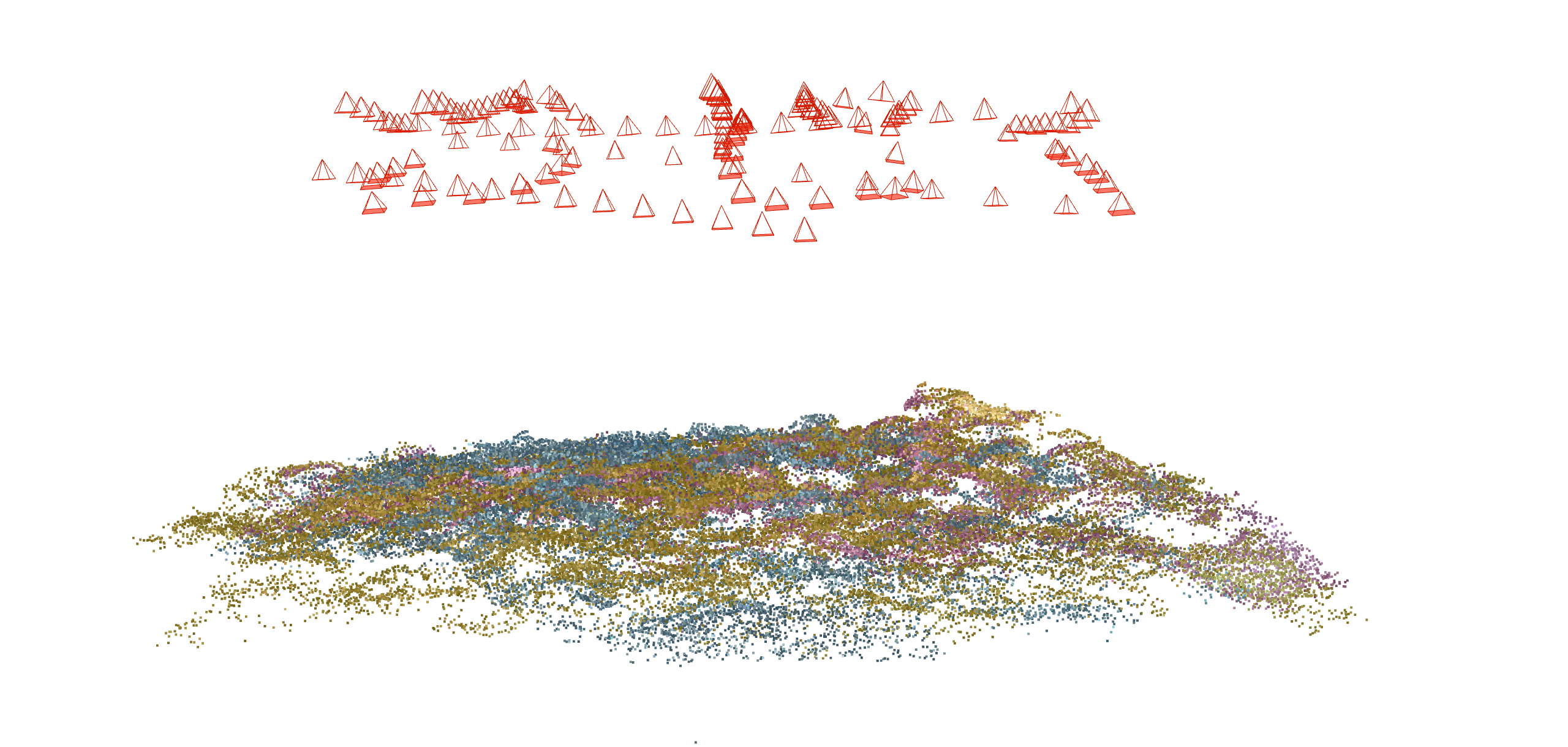} & \cloudfig{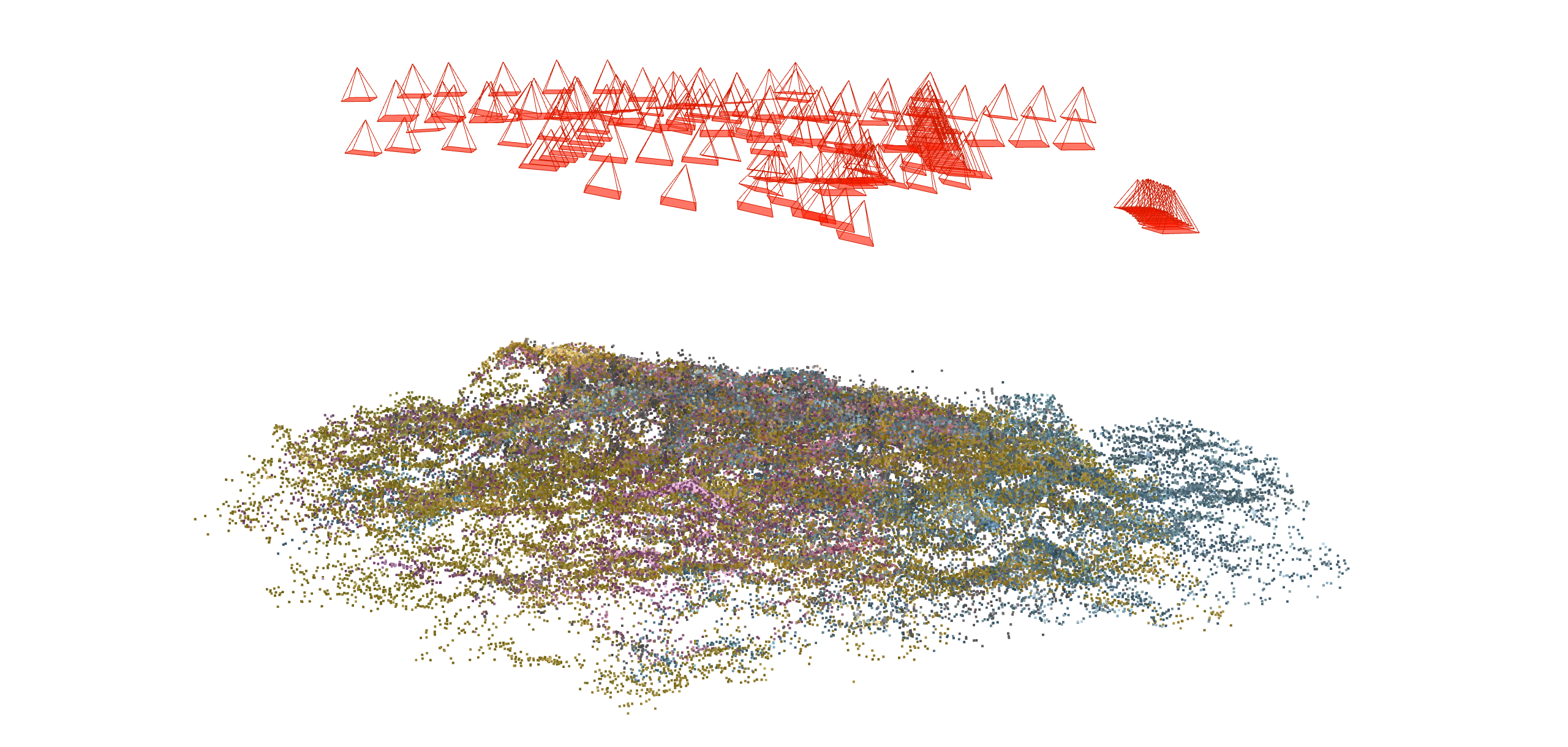}
\\
\pairfig{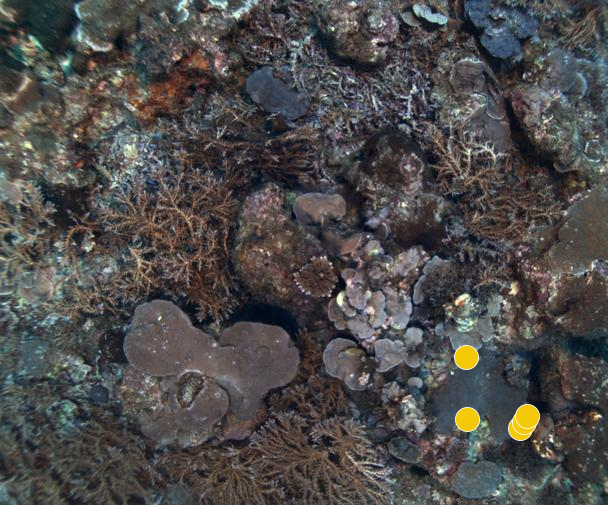} & \pairfig{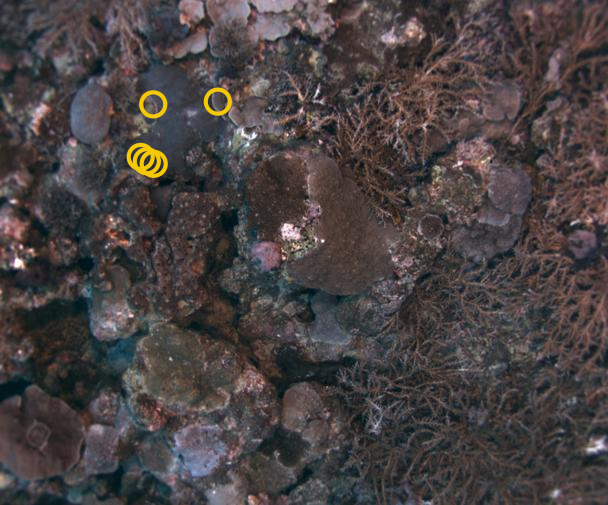} & \pairfig{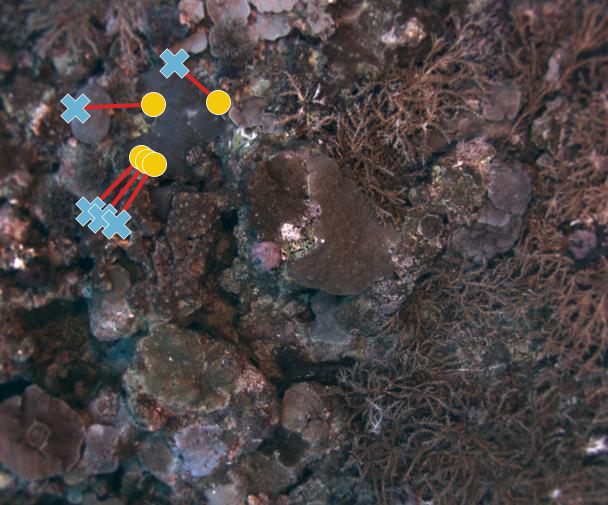} & \pairfig{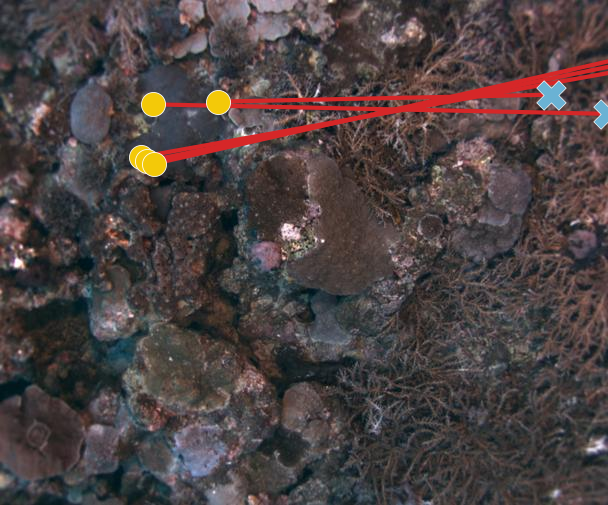} & \pairfig{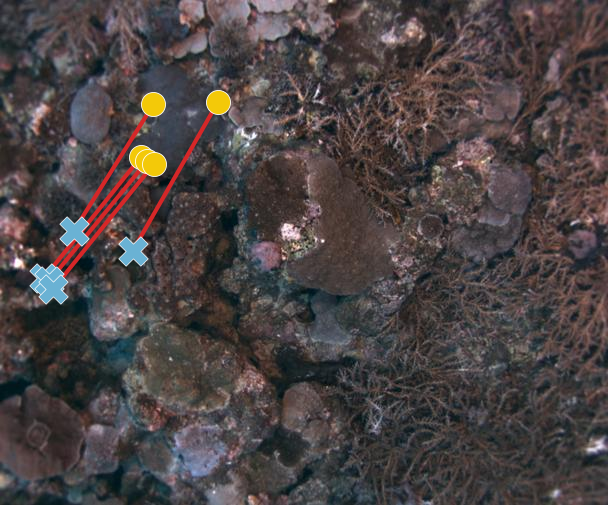}
& \pairfig{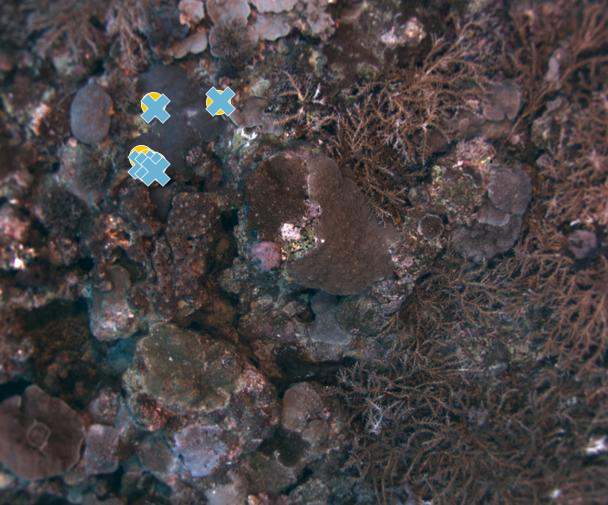}
\\[-0.1cm] %
\includegraphics[width=0.16\textwidth]{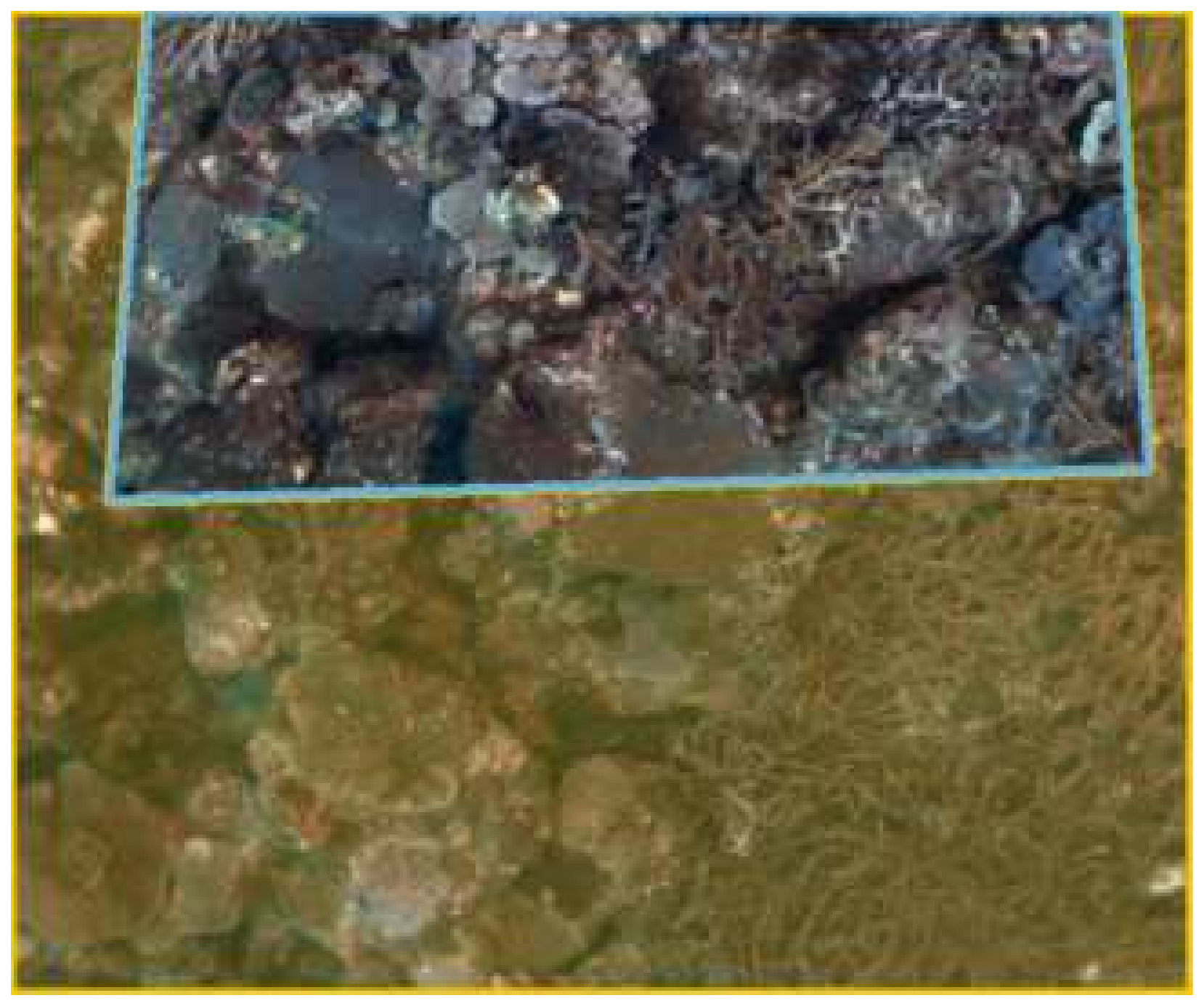}
& \pairfig{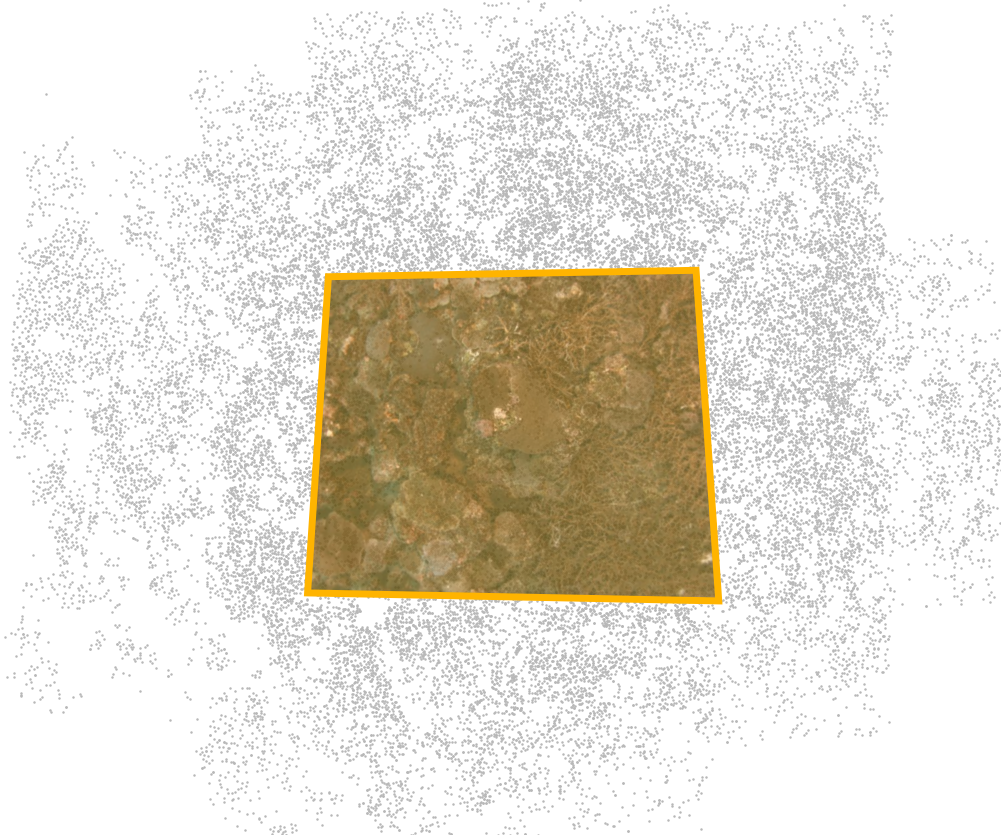} & \pairfig{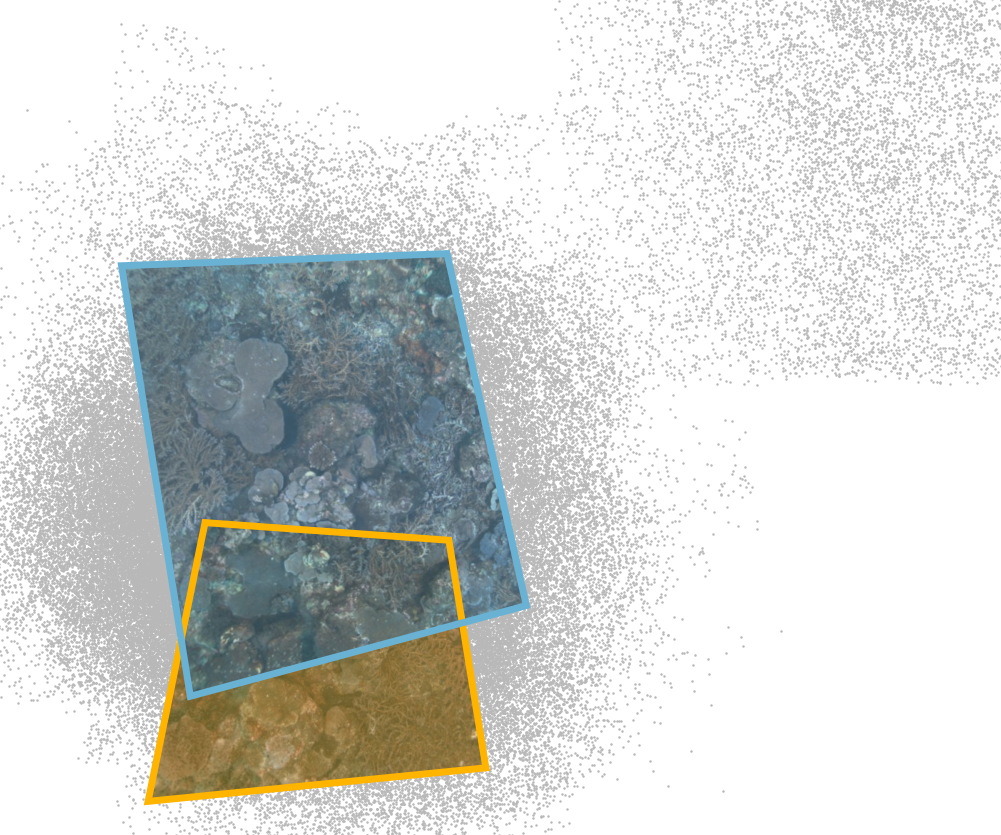} & \pairfig{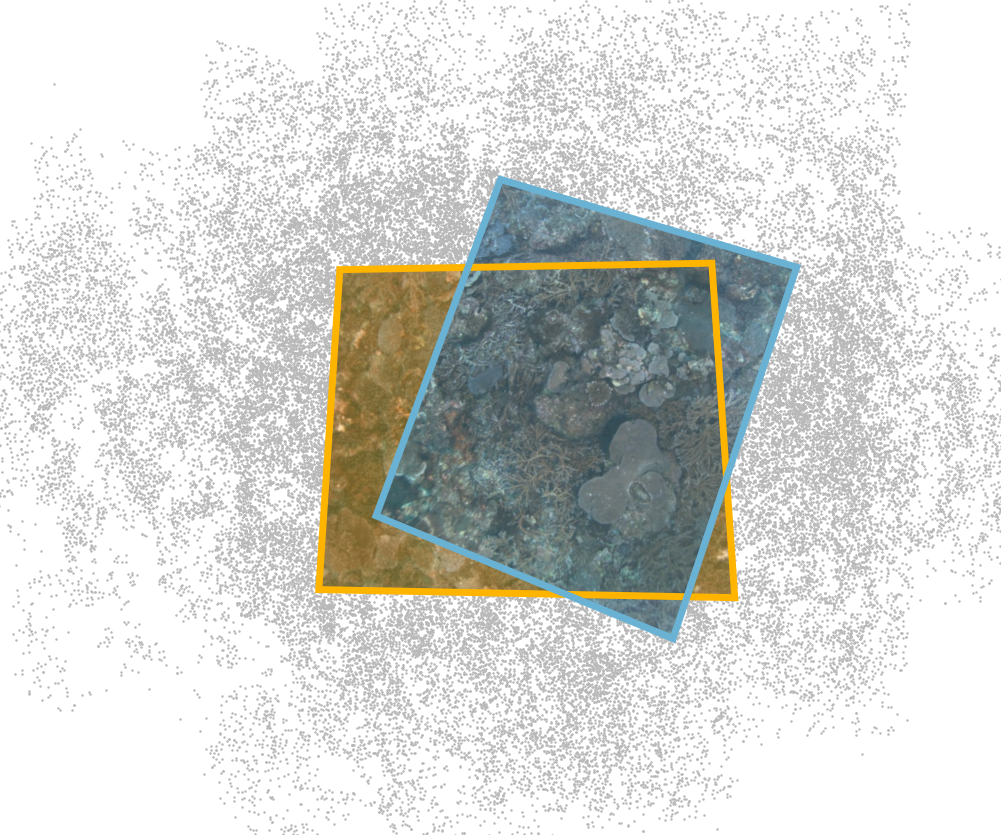} & \pairfig{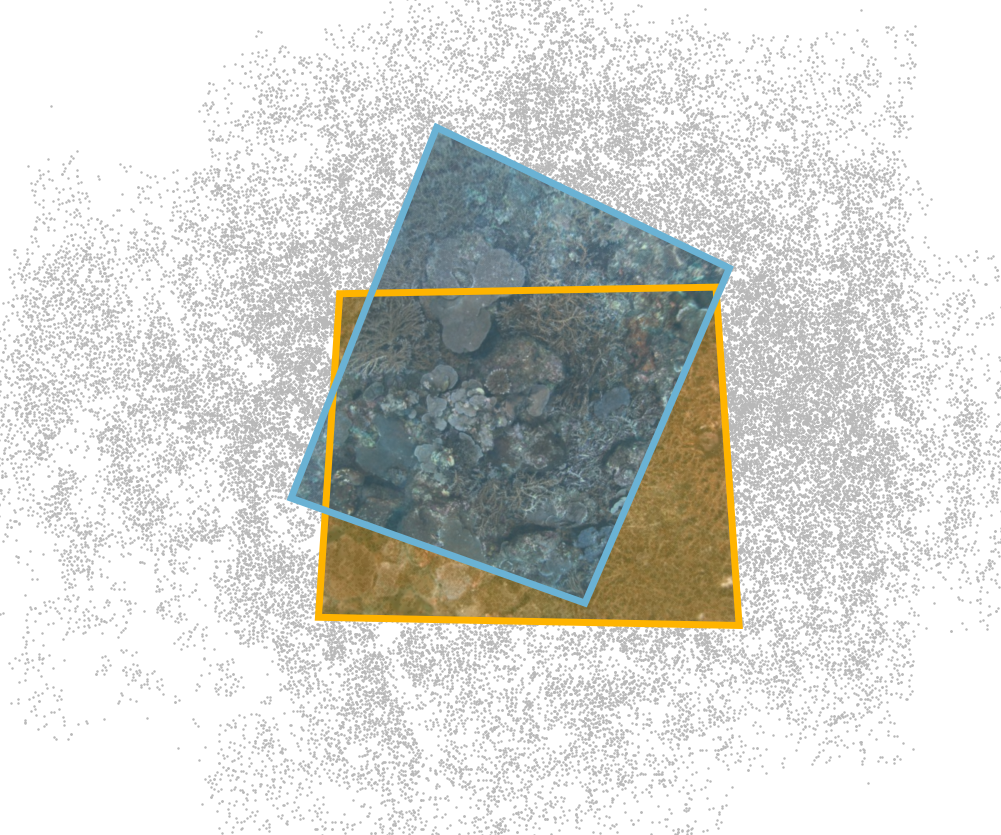}
& \pairfig{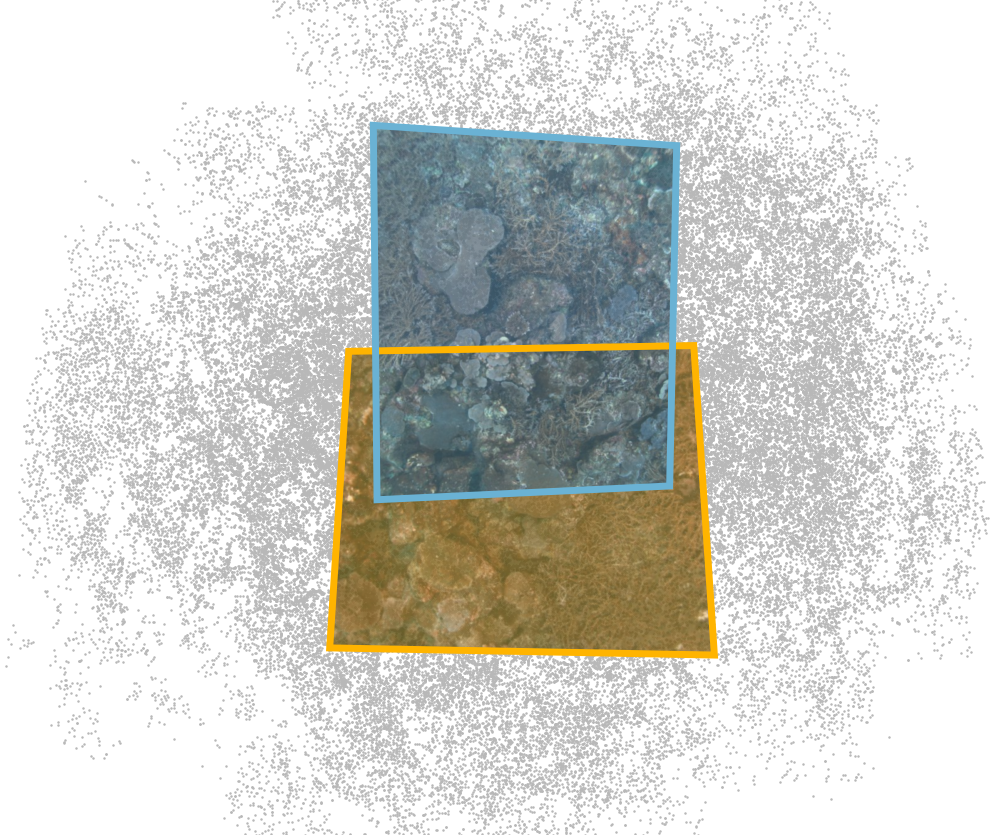}
\\
\end{tabular}
\caption{\textbf{Qualitative comparison on ssk-s01.}
Top: each method's reconstruction, with cameras in red and points colored by session.
Middle: an annotated cross-session image pair (SSK16 query, SSK17 database), with annotated correspondences (yellow circles), points projected by each reconstruction (blue crosses), and reprojection errors (red lines); hollow circles cannot be projected because the query image is not registered.
Bottom: the query warped onto the database image (left), and each reconstruction from above with the query (blue) and database (yellow) image footprints.
COLMAP registers only SSK17 images, MapAnything distorts the scene, and COLMAP + ICP and COLMAP + BUFFER-X misplace the footprints; only our method recovers the expected overlap.}
\label{fig:points3d_images}
\vspace*{-0.3cm}
\end{figure*}

\begin{figure}[t]
\centering
\footnotesize
\begin{tabular}{ccc}
\multicolumn{3}{c}{{\myline{myblue}} Session 1 {\myline{myyellow}} Session 2 {\myline{mypink}} Session 3 {\myline{mygrey}} Multiple sessions}\\
\end{tabular}
\footnotesize
\begin{tabular}{cc}
\includegraphics[height=0.35\linewidth]{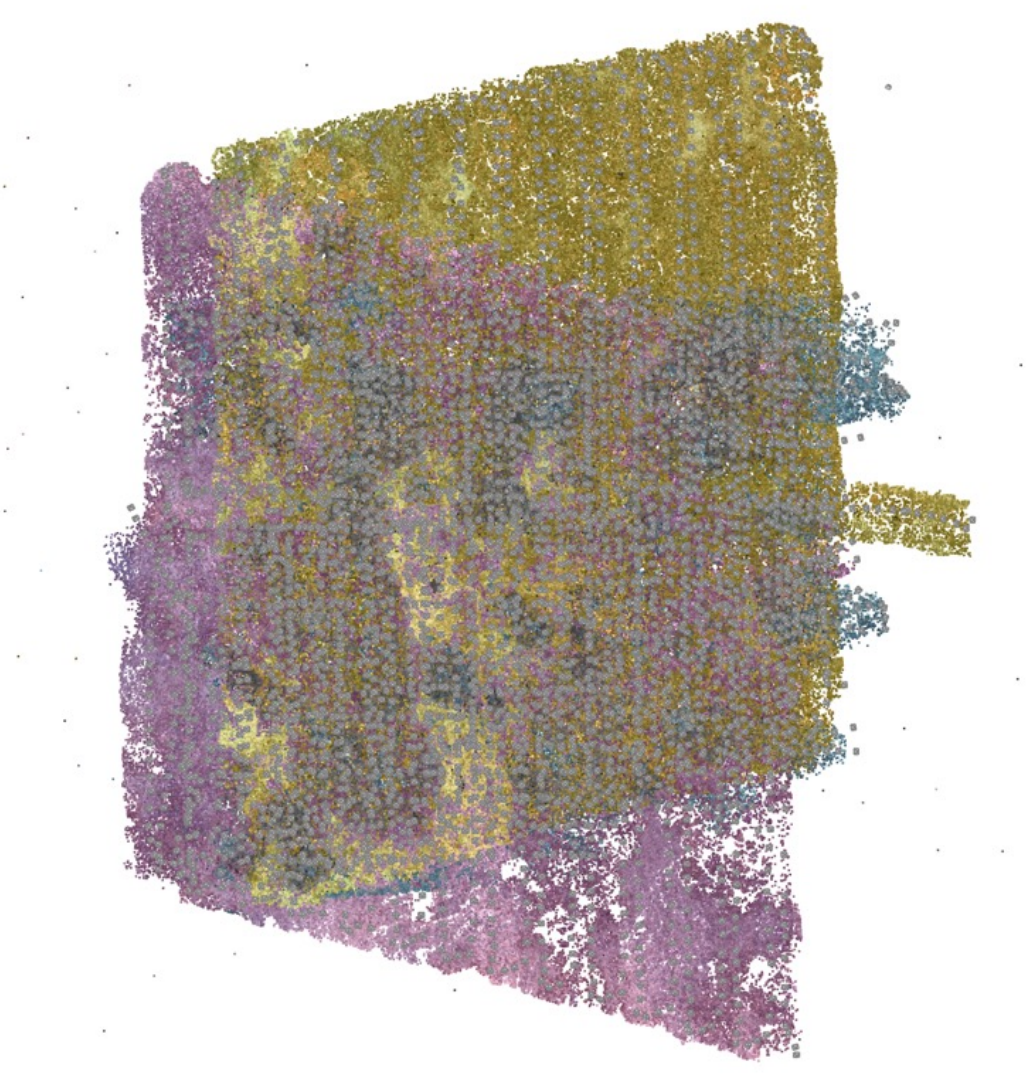} 
&
\includegraphics[height=0.35\linewidth]{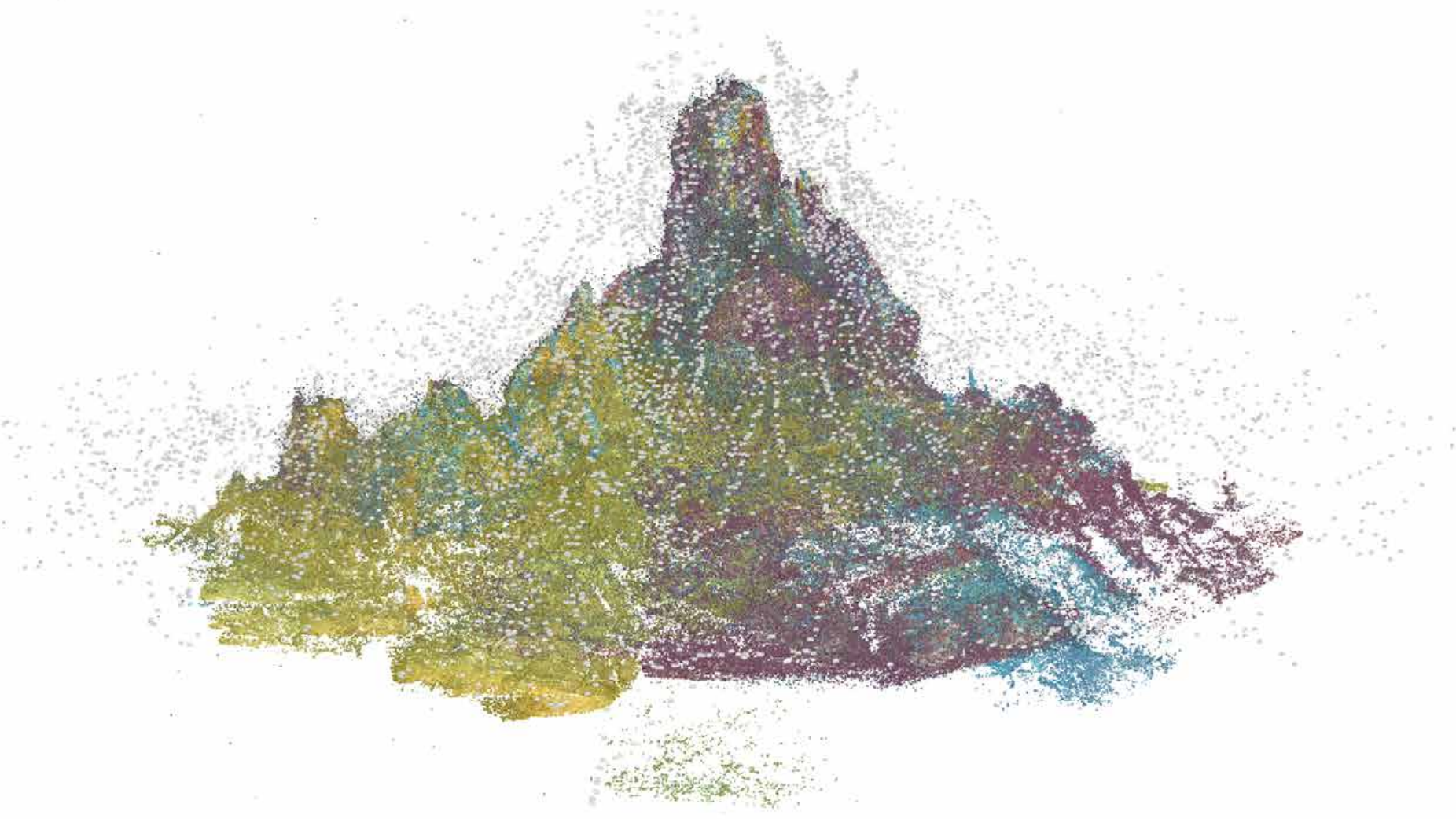} 
\\ (a) ssk-full & (b) eff-full
\end{tabular}
\caption{\textbf{Full multi-session reconstructions.}  Reconstructions of the full Sesoko (left) and Eiffel (right) datasets. Points are colored according to the survey year, showing that observations acquired at different times are jointly reconstructed into a single coherent 3D model despite substantial temporal appearance and structural changes.
}
\label{fig:full_reconstructions}
\end{figure}

\begin{figure}[t]
\centering
\scriptsize
\begin{tabular}{c}
{\myline{myblue}} COLMAP {\myline{myyellow}} COLMAP + ICP {\myline{mypink}} COLMAP + BUFFER-X {\myline{mygreen}} Ours \\
\end{tabular}
\includegraphics[width=0.65\linewidth]{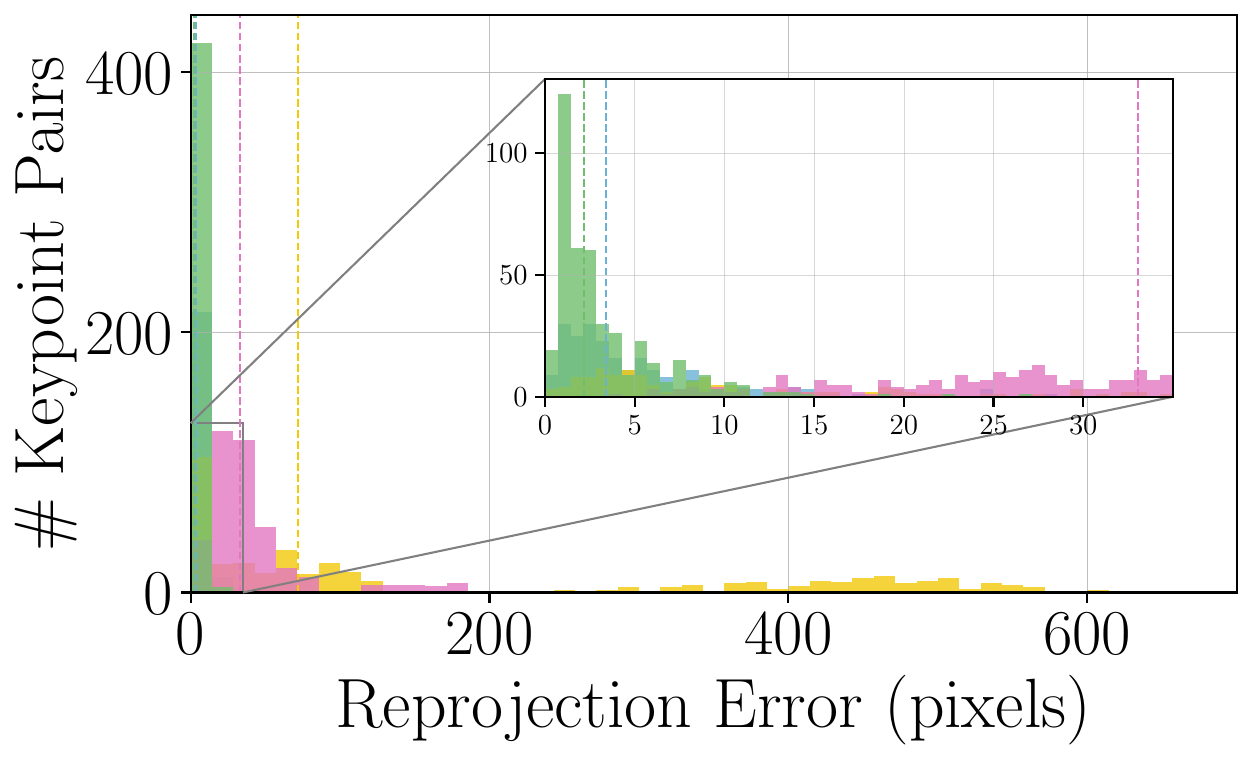}
\vspace*{-0.2cm}
\caption{\textbf{Histogram of reprojection error over the evaluation subsets.}
Distribution of pixel reprojection error for the manually annotated cross-session correspondences of ssk-s01 to ssk-s05, pooled. The inset enlarges the low-error range. Median values for each method are indicated by vertical dashed lines.}
\label{fig:rpe_histograms}
\vspace*{-0.2cm}
\end{figure}

Our method, which enforces cross-session correspondences during reconstruction, registers images in a single model and evaluates every annotated point of all subsets, with median reprojection errors tightly concentrated between 1.36 and 3.94~px across subsets (Table~\ref{tab:main_results}, Figure~\ref{fig:rpe_histograms}). Between 13\% and 25\% of its 3D points are observed in more than one session, indicating that the sessions are connected through shared structure.

Post-hoc ICP and BUFFER-X alignment of the independently reconstructed sessions produce median errors between 40 and 405~px and between 28 and 68~px, respectively, on the subsets. These errors indicate that registering the sessions by their geometry alone leaves them only coarsely aligned.

Aligning the same independent reconstructions with our cross-session correspondences (Sim(3)) instead reaches between 2.40 and 6.60~px across the subsets, supporting our key finding that the establishment of strong cross-session correspondences is what enables long-term multi-session reconstruction under substantial appearance change. Comparing our joint method against Sim(3)~+~BA using our correspondences, we show that enforcing them within the joint reconstruction, rather than after independent reconstructions, is what makes them most effective.

Given only its own SIFT correspondences, standard COLMAP produces fragmented reconstructions (Figure~\ref{fig:points3d_images}), and at most 0.4\% of its 3D points are observed in more than one session (Table~\ref{tab:main_results}). On ssk-s03, COLMAP approaches our accuracy (2.36 vs 2.15~px) on only 51 of the 103 points, but only 0.3\% of its 3D points are shared across sessions, compared with 25\% for our method. %

MapAnything similarly struggles in this long-term, multi-visit setting. Although it aggregates information from multiple sessions, it cannot register the sessions consistently, with median reprojection errors of 77 to 603~px (Table~\ref{tab:main_results}). Processing all images in a single forward pass without selecting or rejecting cross-session pairs, MapAnything absorbs structural and appearance inconsistencies between sessions directly into the reconstruction (Figure~\ref{fig:points3d_images}). On full datasets, MapAnything failed due to excessive memory requirements to process every image in a single forward pass. In contrast, our method integrates data captured years apart without the fragmentation or inconsistency observed in the baseline methods by combining robust intra-session matching with learned cross-session correspondence establishment.

Figure~\ref{fig:points3d_images} compares the resulting reconstructions and pixel-level alignment, demonstrating the need for enforced cross-session correspondences in our framework. We observe that post-hoc alignment based on geometry alone produces coarsely aligned reconstructions that are unsuitable for fine-grained environmental monitoring, while the evaluated joint reconstruction methods do not connect the sessions into a single coherent reconstruction. Figure~\ref{fig:full_reconstructions} shows that our method produces complete and aligned reconstructions across multiple sessions for the full datasets.

Together, these results show that coherent multi-visit SfM under substantial appearance change requires cross-session correspondences established through complementary visual place recognition, learned matching, and joint optimization.

\section{Ablation Studies}
\label{sec:ablations}

\begin{figure}[t]
\centering
\begin{overpic}[width=0.65\linewidth]{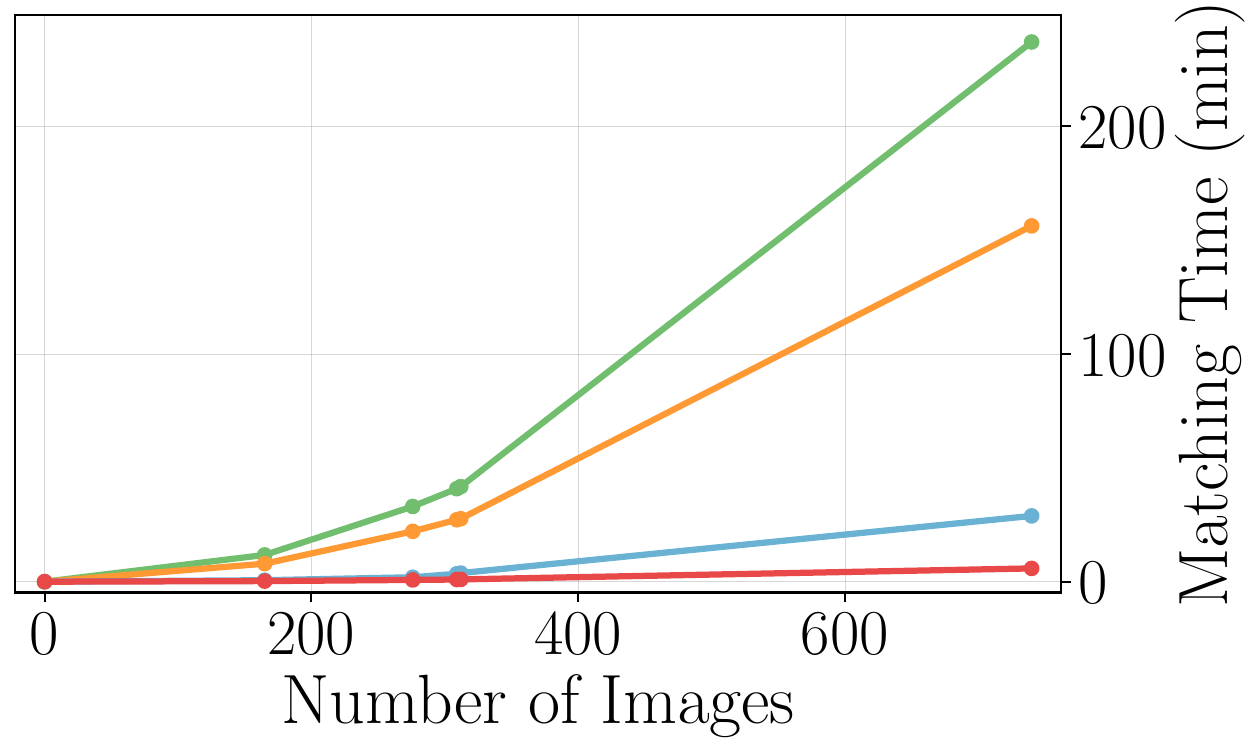}
  \put(7,49){\scriptsize \myline{mygreen} Learned, all pairs}
  \put(7,41.5){\scriptsize \myline{myorange} Learned, cross-session pairs}
  \put(7,34){\scriptsize \myline{myblue} Learned, VPR pairs (Ours)}
  \put(7,26.5){\scriptsize \myline{myred} SIFT only (COLMAP)}
\end{overpic}
\caption{\textbf{Feature matching time against the number of images} on ssk-s01 to ssk-s05 (SIFT exhaustive matching plus learned matching, as in Table~\ref{tab:scalability}). Constraining learned matching to cross-session image pairs, and further to the candidates selected by VPR, substantially reduces matching time compared to exhaustive matching.}
\label{fig:feature_matching_time}
\vspace*{-0.3cm}
\end{figure}

The following experiments examine which features and matcher yield the most effective cross-session correspondences while balancing computational cost (Table~\ref{tab:ablation_features}), how sensitive the pipeline is to the VPR threshold (Table~\ref{tab:ablation_vpr}), and how restricting learned matching to VPR-selected pairs enables scalability to large datasets (Table~\ref{tab:scalability}, Figure~\ref{fig:feature_matching_time}). All experiments use ssk-s05, our largest subset.

\textbf{Cross-session pairs and VPR threshold.} Table~\ref{tab:ablation_features} varies the learned features and matcher used for the cross-session image pairs within our framework. SuperPoint with LightGlue is the most accurate (2.00~px, with all 145 annotated points evaluated). SIFT and ORB are faster (16--38~min vs.\ 50~min) but much less accurate (3.21--12.13~px) and connect few 3D points across sessions (at most 5\% vs.\ our 21\%).  

Table~\ref{tab:ablation_vpr} varies the VPR threshold $\tau$ that selects these pairs. For $\tau \geq 0.50$, the reprojection error remains between 2.00 and 2.09~px on all 145 annotated points, while the 3D points shared across sessions and the runtime increase with the number of candidate pairs. Our method achieves this accuracy while matching 15.6\% of the cross-session image pairs, in approximately a quarter of the runtime required to match all cross-session pairs.

\begin{table}[t]
\centering
\scriptsize
\setlength{\tabcolsep}{2pt}
\caption{\textbf{Ablation of cross-session features and matchers.} All configurations use the same VPR-selected image pairs. SIFT-only adds no learned matches. $^\dagger$: our configuration.}
\begin{tabular}{llcccc}
Features & Matcher &
\makecell{RPE \\ (px) $\downarrow$} &
\makecell{$n$ \\ (/145) $\uparrow$} &
\makecell{Shared 3D pts \\ ($\geq$2 sess.) $\uparrow$} &
\makecell{Time \\ (min) $\downarrow$} \\
\cmidrule(lr){1-2} \cmidrule(lr){3-6}
SuperPoint$^\dagger$ & LightGlue$^\dagger$ & \textbf{2.00} & 145 & 116.8k (21\%) & 50 \\
SuperPoint & BruteForce & 2.26 & 141 & 55.2k (11\%) & 44 \\
ALIKED & LightGlue & 2.28 & 143 & \textbf{267.2k} (37\%) & 66 \\
ALIKED & BruteForce & 2.31 & 143 & 175.2k (27\%) & 74 \\
SIFT & LightGlue & 3.21 & 128 & 3.9k (0.9\%) & 20 \\
SIFT & BruteForce & 5.09 & 128 & 1.1k (0.3\%) & 16 \\
ORB & BruteForce & 12.13 & 129 & 23.2k (5\%) & 38 \\
\cmidrule(lr){1-6}
\multicolumn{2}{l}{SIFT-only} & 5.09 & 128 & 1.1k (0.3\%) & \textbf{12} \\
\end{tabular}
\label{tab:ablation_features}
\vspace*{-0.15cm}
\end{table}

\begin{table}[t]
\centering
\scriptsize
\setlength{\tabcolsep}{3pt}
\caption{\textbf{Sensitivity to the VPR threshold $\tau$.} $\tau = 1.00$ matches all cross-session pairs.}
\begin{tabular}{cccccc}
Threshold $\tau$ &
\makecell{Candidate pairs \\ (\% cross-sess.)} &
\makecell{RPE \\ (px) $\downarrow$} &
\makecell{$n$ \\ (/145) $\uparrow$} &
\makecell{Shared 3D pts \\ ($\geq$2 sess.) $\uparrow$} &
\makecell{Runtime \\ (min) $\downarrow$} \\
\cmidrule(lr){1-1} \cmidrule(lr){2-6}
0.40 & 188 (0.1) & 2.47 & 129 & 12.2k (3\%) & 17 \\
0.45 & 1,711 (1.0) & 2.39 & 132 & 42.2k (9\%) & 19 \\
0.50 & 8,580 (4.8) & 2.09 & 145 & 84.6k (16\%) & 29 \\
0.55 (Ours) & 27,700 (15.6) & \textbf{2.00} & 145 & 116.8k (21\%) & 50 \\
0.60 & 60,729 (34.2) & 2.09 & 145 & 136.9k (23\%) & 81 \\
0.65 & 100,027 (56.3) & 2.05 & 145 & 147.4k (24\%) & 115 \\
0.70 & 133,555 (75.1) & 2.06 & 145 & 153.2k (25\%) & 146 \\
0.80 & 169,010 (95.0) & 2.06 & 145 & 157.6k (25\%) & 178 \\
\midrule
1.00 (all pairs) & 177,836 (100) & 2.05 & 145 & \textbf{158.3k} (25\%) & 186 \\
\end{tabular}
\label{tab:ablation_vpr}
\vspace*{-0.3cm}
\end{table}

\textbf{Computational tractability and scalability.} Table~\ref{tab:scalability} and Figure~\ref{fig:feature_matching_time} show that enforcing cross-session correspondences during reconstruction can be achieved in a computationally tractable manner, supporting the practical applicability of our proposed framework. We evaluate four configurations of increasing selectivity. The first configuration uses SIFT features with exhaustive matching across all images, representing a classical baseline with low computational overhead but limited robustness. The second configuration performs exhaustive SIFT matching and exhaustive learned feature matching across all image pairs. This configuration represents a maximally exhaustive strategy intended to test whether increased matching coverage improves reconstruction quality, albeit at a substantially higher computational cost. The third configuration restricts learned feature matching to cross-session image pairs. The fourth configuration further restricts learned matching to cross-session image pairs identified as likely correspondences via visual place recognition (Ours).

As the number of images increases, exhaustive learned matching incurs a rapidly growing computational cost (Figure~\ref{fig:feature_matching_time}), reaching 277 minutes of total runtime for 740 images (Table~\ref{tab:scalability}). Restricting learned feature matching to cross-session image pairs reduces the total runtime to 186 minutes, while visual place recognition-guided candidate selection further reduces it to 50 minutes, a reduction of \textbf{82\%} relative to exhaustive learned matching.

Importantly, reducing the computational cost of cross-session matching does not reduce reconstruction accuracy. On ssk-s05, learned matching restricted to the visual place recognition candidates achieves a median reprojection error of 2.00~px, while learned matching of all cross-session image pairs yields 2.05~px (Table~\ref{tab:ablation_vpr}) and learned matching of all image pairs 2.14~px. Together, these results show that selective cross-session matching preserves reconstruction accuracy at a fraction of the computational cost, making the proposed framework scalable to larger long-term, multi-visit datasets.

\begin{table}[t]
\centering
\scriptsize
\setlength{\tabcolsep}{3pt}
\caption{\textbf{Computational cost of cross-session matching strategies on ssk-s05 (740 images, 608$\times$505 pixels).}}
\begin{tabular}{lcccc}
\makecell[l]{Learned \\ matching on} &
\makecell{Learned pairs \\ matched} &
\makecell{Matching \\ (min) $\downarrow$} &
\makecell{Total runtime \\ (min) $\downarrow$} &
\makecell{Peak RAM / GPU \\ (GB) $\downarrow$} \\
\cmidrule(lr){1-1} \cmidrule(lr){2-5}
None (SIFT only) & -- & 6 & 12 & 2.9 / 0.9 \\
All pairs & 273,430 & 237 & 277 & 5.8 / 4.6 \\
Cross-session pairs & 177,836 & 156 & 186 & 4.3 / 4.6 \\
VPR pairs (Ours) & 27,700 & \textbf{29} & \textbf{50} & 3.8 / 4.6 \\
\end{tabular}
\label{tab:scalability}
\vspace*{-0.1cm}
\end{table}

\section{Discussion and Conclusion}
\label{sec:conclusion}

We presented a framework for long-term, multi-session 3D reconstruction in underwater environments subject to substantial appearance and structural change. The framework aligns visual surveys captured years apart by enforcing cross-session correspondences directly during Structure-from-Motion optimization, rather than relying on post-hoc alignment of independently reconstructed sessions. %

The results show that when appearance change is severe and stable geometric landmarks are sparse, aligning independently reconstructed sessions by their geometry alone is insufficient. Reliable camera poses and geometry across time instead require cross-session correspondences, and integrating them during a joint optimization gives the most accurate and complete reconstructions. Registering repeated underwater surveys in a shared coordinate frame supports downstream applications including structural change analysis, long-term monitoring, and informed reef intervention planning. Our results further show that this can be achieved without prohibitive computational cost by selectively applying learned feature matching to image pairs identified as likely cross-session matches by visual place recognition.

Reconstruction quality depends on sufficient visual overlap across visits, and rejection of inconsistent correspondences relies on unchanged structure dominating the image pairs that are verified. Alignment therefore degrades in cases such as extensive coral displacement following storm damage, where few stable geometric landmarks may remain to support reliable cross-session alignment. Future work will investigate appearance-invariant representations to improve robustness under extreme long-term change, adaptive correspondence-selection strategies, and explicit refractive modeling for underwater scenes. %

\small
\bibliographystyle{IEEEtran}
\bibliography{references}

\end{document}